\documentclass[lettersize,journal]{IEEEtran}
\usepackage{amsmath,amsfonts}
\usepackage{algorithmic}
\usepackage{algorithm}
\usepackage{array}
\usepackage[caption=false,font=normalsize,labelfont=sf,textfont=sf]{subfig}
\usepackage{textcomp}
\usepackage{float}
\usepackage{url}
\usepackage{verbatim}
\usepackage{graphicx}
\usepackage{cite}
\usepackage{bbold}

\usepackage[colorlinks,bookmarksopen,bookmarksnumbered,citecolor=red,urlcolor=red]{hyperref}
\usepackage{amsmath,amssymb,amsfonts}
\usepackage{graphicx}
\usepackage{multirow}
\usepackage{booktabs}
\usepackage[table,xcdraw]{xcolor}

\definecolor{LightGray}{rgb}{0.9, 0.9, 0.9}
\definecolor{LightBlue}{rgb}{0.68, 0.85, 0.9}
\begin{document}

\title{Multi-Group Equivariant Augmentation for Reinforcement Learning in Robot Manipulation
}

\author{Hongbin Lin, Juan Rojas, and Kwok Wai Samuel Au
        % <-this % stops a space
\thanks{Manuscript received August, 2025;}}

\markboth{Journal of \LaTeX\ Class Files,~Vol.~14, No.~8, August~2025}%
{Shell \MakeLowercase{\textit{et al.}}: A Sample Article Using IEEEtran.cls for IEEE Journals}

\IEEEpubid{0000--0000/00\$00.00~\copyright~2025 IEEE}

\maketitle
%----------------------
\begin{abstract}
%----------------------
Sampling efficiency is critical for deploying visuomotor learning in real-world robotic manipulation. 
While task symmetry has emerged as a promising inductive bias to improve efficiency, most prior work is limited to isometric symmetries---applying the same group transformation to all task objects across all timesteps. 
In this work, we explore non-isometric symmetries, applying multiple independent group transformations across spatial and temporal dimensions to relax these constraints. 
We introduce a novel formulation of the partially observable Markov decision process (POMDP) that incorporates the non-isometric symmetry structures, and propose a simple yet effective data augmentation method, \textbf{M}ulti-Group \textbf{E}quivariance \textbf{A}ugmentation (MEA). 
We integrate MEA with offline reinforcement learning to enhance sampling efficiency, and introduce a voxel-based visual representation that preserves translational equivariance. 
Extensive simulation and real-robot experiments across two manipulation domains demonstrate the effectiveness of our approach.

\end{abstract}
%----------------------
\begin{IEEEkeywords}
Reinforcement Learning, Symmetry, and Robot Manipulation.
\end{IEEEkeywords}
%----------------------
\section{INTRODUCTION}

Visuomotor learning has gained significant attention in robotic manipulation due to its potential to improve generalization, robustness, and long-term reasoning \cite{kalashnikov2018scalable, levine2016end}. 
However, directly learning visuomotor policies on real robots, also known as on-robot learning, presents challenges due to low sampling efficiency. 
For imitation learning, tens of demonstrations are often required (e.g., 50 demonstrations for fine-grained robot manipulation tasks \cite{zhao2023learning}), while reinforcement learning (RL) demands even larger datasets, such as 10 hours of data for robot grasping tasks \cite{wu2023daydreamer}. 
The sampling efficiency in RL can be enhanced by incorporating demonstrations, as shown by SERL \cite{luo2024serl}, which learns object relocations with only 10 demonstrations and 105 minutes of data. 
Nonetheless, the need to collect substantial data from real robots significantly increases both the cost of human labor and the operational time of the robot.
Transfer learning offers an alternative to alleviating the need for large-scale, domain-specific real-robot data. 
It involves sim-to-real transfer \cite{singh2024dextrah}, training visuomotor policies in simulation before deploying them on real robots, and pre-training \cite{nair2022r3m}, using large-scale, out-of-domain data prior to policy transfer.
However, these methods often incur higher computational costs and, in many cases, still require significant real-robot data for fine-tuning \cite{kim2024openvla}.

\begin{figure}[!tbp]
  \centering
  \includegraphics[width=1\hsize]{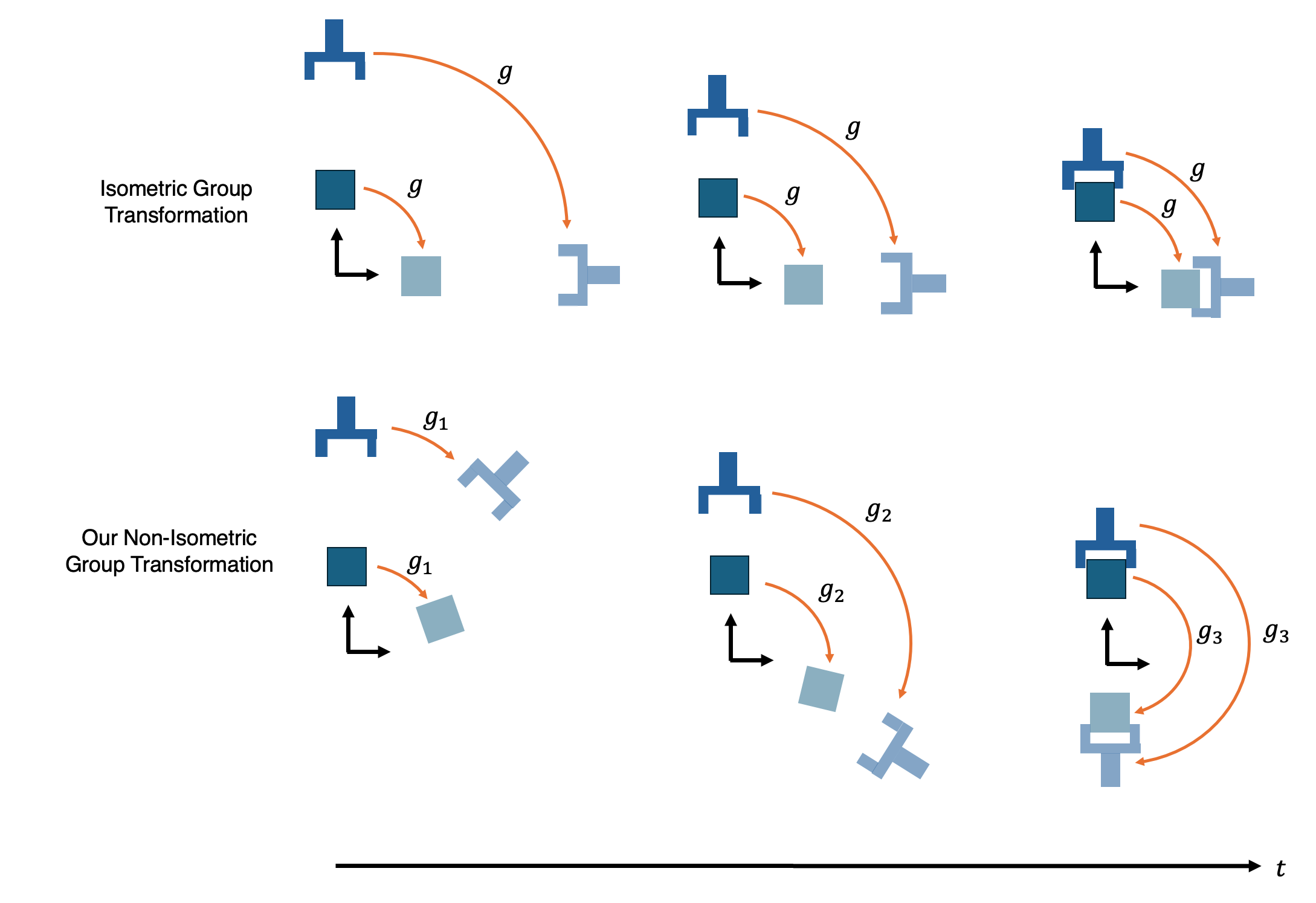}
  \caption{Isometric group transformation vs. our non-isometric group transformation. 
 On a 2D plane, a gripper grasps a square object over 3 timesteps. 
We compare augmented trajectories (light blue) based on the ground-truth trajectory (dark blue). 
In the isometric case, both gripper and object are rotated by the same group element $g$ at all timesteps, preserving their relative pose. 
In contrast, our non-isometric transformation uses independent group elements $g_1$, $g_2$, and $g_3$ across timesteps, applying distinct non-isometric group representations to the gripper and object at each step.
This increases symmetry variance of augmentation by altering relative spatial relationships between the gripper and the target object, enhancing sampling efficiency.}
  \vspace{-0.45cm}
 \label{fig:iso_vs_noniso}
\end{figure}

A promising approach to improving the sampling efficiency of visuomotor learning is to exploit intrinsic symmetries in robotic manipulation as task-specific priors \cite{lin2020invariant, zeng2021transporter}. 
For instance, prior work has studied isometric pick symmetry in robot-picking tasks: when a target picking object is rotated in the plane, the optimal grasp pose must also rotate with the same group transform. 
By leveraging isometric rotation-equivariance, this symmetry can be used with data augmentation \cite{laskin2020curl, kostrikov2020image} and equivariant models \cite{wang2022equivariant, nguyen2023equivariant} to enhance sampling efficiency. 
However, the isometric group transformation remains fixed between task objects (e.g., the transformation between the gripper and the target object) and across timesteps within a visuomotor trajectory (see Fig. \ref{fig:iso_vs_noniso}). 
The limited variance in both spatial and time dimensions restricts further improvements in sampling efficiency.
\IEEEpubidadjcol

Few studies explore non-isometric group symmetries, relaxing the isometric assumption to improve sampling efficiency. 
Non-isometric group transformations between task objects have been used as inductive priors in neural descriptor fields \cite{simeonov2022neural, simeonov2023se, chun2023local, ryu2022equivariant}, shape wrapping \cite{biza2023one}, and generative flow models \cite{huang2024imagination}. 
However, these approaches are limited to single-step decision-making formulations. 
Some works have applied non-isometric group transformations in two-step decision-making, such as in pick-and-place tasks, where two independent group transformations are used during both the picking and placing steps \cite{huang2024leveraging, huang2024fourier}. 
Yet, it remains unclear how to extend non-isometric group transformations to multi-step decision-making (i.e., more than two steps).

\begin{figure*}[!tbp]
  \centering
  \includegraphics[width=1.0\hsize]{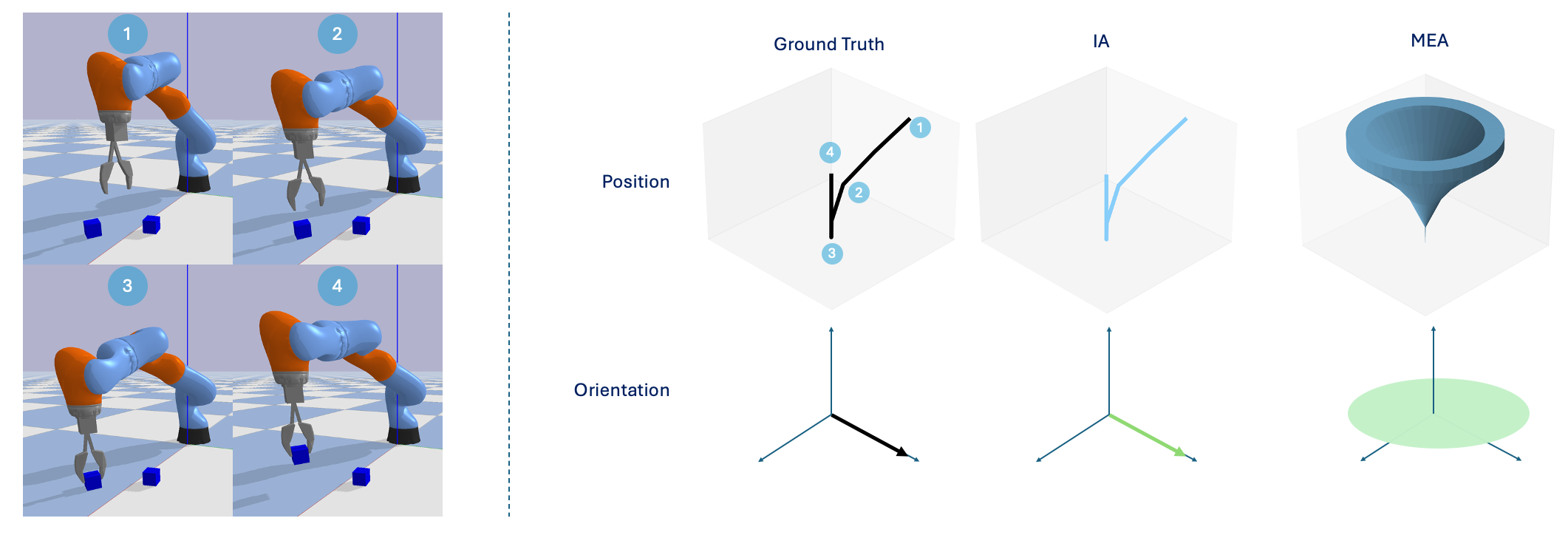}
  \caption{Isometric augmentation (IA) vs. multi-group equivariant augmentation (MEA). In the block-picking task, a robotic manipulator with a gripper is controlled to grasp a block on the floor (left). On the right, we visualize the spanning Cartesian spaces of both gripper’s position and orientation relative to the block within an episode of ground-truth demonstration trajectory. The spanning spaces from left to right are the ground-truth space, the IA space, and our MEA space.}
  \vspace{-0.45cm}
 \label{fig:mea_space}
\end{figure*}

In this paper, we aim to investigate the unaddressed challenges of leveraging group symmetries in visuomotor learning for robotic manipulation. 
Our first goal is to formulate non-isometric group transformations across both spatial and temporal dimensions in the context of multi-step decision making. 
Second, we investigate how multi-group non-isometric symmetries can be exploited to enhance the sampling efficiency of visuomotor policy learning in robot manipulation. 
To this end, we introduce a novel formulation with the partially observable Markov decision process (POMDP) that incorporates symmetry structures, and propose a simple yet effective data augmentation strategy, \textbf{M}ulti-Group \textbf{E}quivariance \textbf{A}ugmentation (MEA), as shown in Fig. \ref{fig:iso_vs_noniso}. 
We further integrate this augmentation with offline RL methods to boost sampling efficiency. 
Additionally, we propose a voxel-based visual representation for visuomotor policies that preserves translational equivariance.

In summary, our main contributions are:
\begin{enumerate}
    \item A novel formulation of non-isometric symmetry within the POMDP framework. 
    We apply multiple independent groups to a trajectory, resulting in non-isometric group transformations for task objects across all timesteps. 
    Our formulation enhances the variance of symmetry in both spatial and temporal dimensions compared to traditional isometric transformations.
    
    \item A novel data augmentation method based on our non-isometric symmetry formulation for visuomotor learning in robot manipulation. 
    Our method ensures valid augmentations even under imperfect symmetries in robotic manipulation through multi-phase augmentation. 
    Additionally, it increases augmentation diversity and relevance by incorporating structured action. 
    This augmentation is integrated with offline RL methods, including a model-free isometric-equivariant RL approach and a model-based non-equivariant RL approach.

    \item A novel equivariant image representation using voxel-based orthogonal projection, which preserves translational equivariance for improved visuomotor policy learning.

    \item Extensive simulation and real-robot experiments across two robot manipulation domains: general manipulation and surgical grasping. 
    Our non-isometric approach achieves over $25\%$ improvement in convergence performance over the isometric-equivariant baseline in general manipulation, and a fourfold improvement over the non-equivariant baseline. 
    Additionally, it reduces the need for demonstration data by over $97.5\%$ in both domains and cuts total training steps and time by over $58\%$ in surgical grasping. 
    Our approach shows no performance degradation in simulation or sim-to-real transfer, while improving sampling efficiency.
\end{enumerate}

\section{Related Works}
\subsection{Symmetry As Inductive Bias in Robot Manipulation}
Researchers leverage symmetry priors as inductive bias in the fields of imitation learning \cite{simeonov2022neural,simeonov2022neural,chun2023local,huang2023edge,lin2023mira,zeng2021transporter,huang2022equivariant,huang2024leveraging,huang2024fourier,ryu2022equivariant,biza2023one,gao2024riemann,ryu2024diffusion,huang2024imagination,zhu2022sample} and reinforcement learning \cite{wang2022equivariant,wang2022mathrm,wang2022robot,nguyen2023equivariant,lin2020invariant} for robot manipulation.
Among these works, isometric symmetry has been widely studied, where the isometric symmetry can be used as inductive bias with data augmentation \cite{lin2020invariant}, image representation \cite{lin2023mira}, or equivariant neural networks---including convolutional neural networks \cite{zhu2022sample,wang2022equivariant,wang2022mathrm,wang2022robot,nguyen2023equivariant}, graph-based networks \cite{huang2023edge}, and diffusion networks \cite{ryu2024diffusion}.
These works are limited by their isometric transformation of a single group on task objects across the entire task, which limits their performance due to the lack of diversity.
Some works study non-isometric transformation acting on local task objects in the grasping tasks using neural descriptor field \cite{simeonov2022neural,simeonov2023se,chun2023local,ryu2022equivariant}, shape wrapping \cite{biza2023one}, generative flow models \cite{huang2024imagination}.
The non-isometric transform for local task object is also extended to pick-and-place tasks \cite{huang2022equivariant,huang2024leveraging,huang2024fourier,gao2024riemann,huang2024imagination}, where non-isometric transformations are individually applied to the target object's and goal location's poses with two independent groups, ensuring the preservation of symmetry properties.
Compared to their works, we extend the non-isometric transformation to multi-step decision making, where non-isometric transformations are applied across entire timesteps and across all task objects.

\subsection{Data Augmentation in Visual Reinforcement Learning}
Data augmentation in visual reinforcement learning has been widely explored in the literature---see comprehensive reviews in \cite{ma2022comprehensive,hu2024revisiting}.
Domain randomization \cite{sadeghi2016cad2rl,tobin2017domain,pinto2017asymmetric} is used to enhance the generalization in sim-to-real transfer, where simulation properties, such as textures, lighting, and camera location, are randomized to increase the data diversity.
Compared to domain randomization, our method generates new trajectories using the original data without collecting additional data from the environment. 
Researchers explore image processing for data augmentation, including randomly cropping and shifting images \cite{yarats2021image,laskin2020reinforcement,laskin2020curl}, random convolution \cite{lee2019network,hansen2021generalization}, and cutout \cite{cobbe2019quantifying}.
For robot manipulation, image properties such as brightness, contrast, and saturation are randomized to enhance generalization and robustness \cite{kalashnikov2018scalable}.
Zeng et al apply random SE(2) transformations to the observed image and use orthogonal projection for equivariant image representation \cite{zeng2021transporter}.
However, none of the previous works investigate non-isometric transformations for multi-step decision making.
Some works focus on generating representations that are equivariant with respect to an object's translation \cite{lin2023mira,zeng2021transporter,zeng2017multi}.
Zeng et al. directly project point clouds to equivariant images using orthogonal projection \cite{zeng2017multi,zeng2021transporter}.
Compared to their works, our image representation---based on a novel voxel-based projection---produces images with fewer artifacts.
Lin et al. generate an equivariant radiance field using orthogonal projection \cite{lin2023mira}.
In contrast, we generate re-projected images for the visuomotor policy's input.

\section{Background}
\subsection{Partially Observable Markov Decision Process}
Decision-making problems can be formalized with \textit{partially observable markov decision process} (POMDP), denoted as a 7-tuple $(S, A, R, T, O,\gamma, \Omega)$, where $S$ is a set of possible states in the environment; 
$A$ is a set of actions that agents perform;
$R(s_t, a_t): S\times A \to \mathbb{R}$ is a reward function; 
$T(s_t,a_t,s_{t+1})$ is the function of transition probability which represents the likelihood of reaching $s_{t+1}$ after taking action  $a_t$ in $s_t$; 
$O(a_{t-1},s_{t},o_{t})$ is observation probability which represents likelihood of observation $o_{t}$ in reached state $s_{t}$ after taking action $a_{t-1}$;
$\gamma \in [0,1]$ is the discount factor; 
and $\Omega$ is the set of observations. 
The goal is to maximize the agent's expected cumulative reward given its policy $\pi$, denoted as 
$\mathbb{E}_{a_t\sim\pi,s_0\sim b_0}\big[ \sum_{t=0}^{T} \gamma^t R(s_t,a_t)\big]$, where $b_0$ is the belief function for the initial state $s_0$, representing the uncertainty of the state. 
In a rollout, the timestep $t$ is within the range of $[0,H]$, where $H$ is the horizon for the terminated timestep. 
We also use a binary flag $\lambda_t\in\mathbb{1}$ to represent if the current state is terminated.
For offline RL, the transitions $(a_t,r_t,o_t,\lambda_t)$ for timestep $t$ in a rollout is assigned to a replay data buffer $\mathcal{D}$ for further offline training as
\begin{equation}
\label{equ:databuffer_assign}
    \mathcal{D} \leftarrow (a_t,r_t,o_t,\lambda_t).
\end{equation}

%----------------------
\subsection{Group-Invariant POMDPs}
%----------------------
We consider a group $G$ acting on a set $X=\{x\}$. The \textit{group action} is denoted as $gx$. Given a mapping $\phi: X \to Y$ and a group $G$ that acts on sets $X$ and $Y$, then $\phi$ is \textit{group-equivariant} if $\phi(g  x)=g  \phi(x)$ and \textit{group-invariant} if $\phi(gx)=\phi(x)$. Following the definition in \cite{nguyen2023equivariant}, a POMDP is \textit{group-invariant} with respect to a isometric group $G$ if it satisfies the following four conditions for $\forall g \in G$:
%----------------------

%----------------------
\begin{subequations}
\begin{align}
    T(s_t,a_t,s_{t+1}) &= T(g s_t,g a_t, g s_{t+1}) \label{equ:condition_GPOMDP_T}, \\
    R(s_t,a_t)&= R(g s_t, g a_t) \label{equ:condition_GPOMDP_R}, \\
    O(a_{t-1},s_{t},o_{t})&= O(g a_{t-1},g s_t,g o_t) \label{equ:condition_GPOMDP_O}, \\
    b_0(s_0) &= b_0(g s_0)\label{equ:condition_GPOMDP_B}.
\end{align}
\end{subequations}
%----------------------
Let $h_t=(o_0,a_0,..,o_{t-1},a_{t-1},o_{t})$ be the history and  $g  h_t = (g o_0, g a_0,.., g o_{t-1},g a_{t-1}, g o_{t})$ be the group $g$ acting on this history. Group-invariant POMDP exhibits nice properties, where 
%----------------------
\begin{enumerate}
    \item its optimal value function $V^{*}$ is group-invariant: $V^{*}(h_t)=V^{*}(g h_t)$;
    \item its optimal Q-function $Q^{*}$ is also group-invariant: $Q^{*}(h_t,a_t) = Q^{*}(g h_t,g a_t)$;
    \item there is at least one group-equivariant deterministic optimal policy $\pi^{*}(gh_t) = g \pi^{*}(h_t)$.
\end{enumerate}
%----------------------
Detailed formulations and proofs can be found in \cite{nguyen2023equivariant}.

%----------------------
\section{Problem Formulation}
%----------------------
\subsection{Multi-Group-Invariant POMDP}
%----------------------
The formulation of group-invariant POMDP uses the isometric group transformation $g$ across all timesteps in a rollout, which limits the variation of transformations.
To address this limitation, we extend the formulation by incorporating multiple independent groups. 
In the following, we elaborate on the POMDP formulation with multiple independent groups that preserves the symmetries.

\noindent{\bf{Definition 1:}} 
%----------------------
Let $g_0,g_1,...,g_H$ be the mutually independent group elements for the timestep $0,1,...,H$ in a rollout, each belonging to their corresponding group spaces $G_0, G_1,...,G_H$, respectively.
A set $\boldsymbol{G_H} = \{G_t\}_{t=0}^H$ holds all the independent group spaces.
%----------------------
We define a POMDP as \textit{multi-group-invariant POMDP} with respect to $\boldsymbol{G_H}$ if it satisfies the following four conditions: for $\forall g_t\in G_t, t\in [0,H] $, we have
%----------------------
\begin{subequations}
    \begin{align}
        T(s_t,a_t,s_{t+1}) &= T(g_t s_t,g_t a_t,g_{t+1} s_{t+1}) \label{equ:condition_noniso_GPOMDP_T}, \\
        R(s_t,a_t) &= R(g_t s_t, g_t a_t) \label{equ:condition_noniso_GPOMDP_R}, \\
        O(a_{t-1},s_{t},o_{t}) &= O(g_{t-1} a_{t-1},g_{t}s_{t},g_{t}o_{t})\label{equ:condition_noniso_GPOMDP_O}, \\
        b_0(s_0) &= b_0(g_0 s_0)
    \label{equ:condition_noniso_GPOMDP_B}.
    \end{align}
\end{subequations}
%----------------------
% \noindent{\bf{Theorem 1:}}  
% %----------------------
% If a POMDP is i) multi-group-invariant with respect to $\boldsymbol{G_H}$, ii) group-invariant with respect to $G$, and iii) $ \forall G_t \in \boldsymbol{G_H}: G \subseteq G_t$, then the POMDP is multi-group-invariant with respect to $\boldsymbol{G_H'}=\{G_t'\}_{t=0}^H$, where $\forall g_t'\in G_t', g\in G,t\in [0,H]: g'_t = g_t g$.
% \newline \newline
%----------------------
% \noindent{\bf{Theorem 2:}} 
% Similar to the formulation of group-invariant POMDP. We define the history at timestep $t$ as $h_t=(o_0,a_0,..,o_{t-1},a_{t-1},o_{t})$. Then, the transformation of multiple independent groups acting on the history $\boldsymbol{g}  h_t$ is formulated as 
% %----------------------
% \begin{equation}
%     \boldsymbol{g}  h_t = (g_0 o_0, g_1  a_1,.., g_{t-1} o_{t-1},g_{t-1} a_{t-1}, g_{t} o_{t})
% \end{equation}
% %----------------------
% If POMDP is multi-group-invariant, then $\forall t\in[0,H]$ we have
% %----------------------
% \begin{enumerate}
%     \item its optimal value function $V^{*}$ is group-invariant: $V^{*}(h_t)=V^{*}(\boldsymbol{g} h_t)$;
%     \item its optimal Q-function $Q^{*}$ is also group-invariant: $Q^{*}(h_t,a_t) = Q^{*}(\boldsymbol{g} h_t,g_t a_t)$;
%     \item there is at least one group-equivariant deterministic optimal policy $\pi^{*}(\boldsymbol{g}  h_t) = g_t \pi^{*}(h_t)$.
% \end{enumerate}
% %----------------------
% We provide the proofs for Theorem 1 and Theorem 2 in Appendix \ref{sec: appendix_proof_1} and \ref{sec: appendix_proof_2}, respectively.

%----------------------
\subsection{Group Representation}
%----------------------
Group representation is a special case of group action, where a vector $x\in\mathbb{R}^d$ is linearly transformed by a group $g$. 
In particular, each group element $g\in G$ maps to an invertible matrix $\rho(g)\in\mathbb{R}^{d\times d}$.
The group representation on $x$ is obtained through the multiplication of the representation matrix $\rho(g)$ and $x$ itself, denoted as $\rho(g)x$.

In this paper, 
% we leverage group representations as group actions on the space $\mathbb{R}^3 \times \mathrm{SO}(2)$, where $\mathrm{SO}(2)$ denotes the special orthogonal group of degree 2.
% %
% We decompose the group element $g\in \mathbb{R}^3\times \mathrm{SO}(2)$ into a tuple $g=(g_l,g_{\theta})$, where $g_l$ and $g_\theta$ are two group elements in $R^3$ and $\mathrm{SO}(2)$, respectively. 
% %
% In the following, 
we introduce three group representations used in our paper: i) trivial representation, ii) translational standard representation, and iii) rotational standard representation.

%----------------------
\subsubsection{Trivial Representation} 
%----------------------
Let $x\in \mathbb{R}^n$ be a vector in an arbitrary $n$-dimensional space and $g$ be a group element, then the trivial representation leaves the vector unchanged as $\rho_0(g)x=x$.
For the simplicity of notation, we denote the trivial representation as $\rho_0x$ for the rest of this paper.
%----------------------
\subsubsection{Translational Standard Representation} 
%----------------------
Let $x\in \mathbb{R}^n$ be a $n$-dimensional vector and $g_l\in \mathbb{R}^n$ is a $n$-dimensional group element, then the translational standard representation transforms $x$ by adding an offset $g_l$ as
\begin{equation}
    \rho_l(g_l)=x+g_l.
\end{equation}

%----------------------
\subsubsection{Rotational Standard Representation} 
The rotational standard representation transforms a vector $x$ differently depending on the vector's space dimension.
%----------------------
%
Let $g_\theta\in \mathrm{SO}(2)$ be a group element.
When $x\in \mathbb{R}$ is a variable in $1$-dimensional space, the rotational standard representation transforms $x$ by adding an offset of $g_\theta$ as
%----------------------
\begin{equation}
    \rho_{\theta}(g_\theta)x = x + g_\theta.
\end{equation}
%----------------------
When $x\in \mathbb{R}^3$ is a variable in $3$-dimensional space, the rotational standard representation transforms $x$ by rotating it around the $z$-axis of the reference frame by an angle of $g_\theta$ as
%----------------------
\begin{equation}
     \rho_\theta(g_\theta)x = 
     \begin{bmatrix}
        \cos (g_\theta) & -\sin (g_\theta) & 0\\
        \sin (g_\theta) & \cos (g_\theta) & 0 \\
        0 & 0 & 1
    \end{bmatrix}x.
\end{equation}

%----------------------
\subsection{Perfect and Imperfect Symmetries}
%----------------------
% {\color{blue}provide an overview here. don't go straight into definition 3.}
% %----------------------
% {\color{blue} this paragraph needs a lot of improvement in written clarity. I will try to improve some.}
%----------------------
To leverage the inductive prior of multi-group-invariant POMDP, one has to find proper groups for all timesteps $g_0,g_1,...,g_H$ and their corresponding group representation that satisfy the multi-group-invariant conditions. 
While transitions transformed by non-trivial representations---such as standard translation and rotation---tend to exhibit greater variance than those under the trivial representation, there may exist some states in a POMDP for which no non-trivial representations satisfy the multi-group invariance conditions.
In the following, we provide a definition for classifying the existence of non-trivial representations for all states in a POMDP that preserves the symmetry properties.

\noindent{\bf{Definition 3:}}  
%----------------------
We say that a POMDP is \textit{perfect symmetry} if there exist non-trivial representations for all $g_t\in G_t$ such that satisfy the multi-group-invariant conditions (\ref{equ:condition_noniso_GPOMDP_T},\ref{equ:condition_noniso_GPOMDP_R},\ref{equ:condition_noniso_GPOMDP_O},\ref{equ:condition_noniso_GPOMDP_B}).
Otherwise, the POMDP is \textit{imperfect symmetry}. 
In a POMDP with imperfect symmetry, we define a state $s_t$ as an \textit{imperfect state} if it can only be transformed by the trivial representation.

%----------------------
\subsection{POMDP for Robot Manipulation}
\label{sec:POMDP in Robot Manipulation}
\begin{figure}[!tbp]
  \centering
  \includegraphics[width=1\hsize]{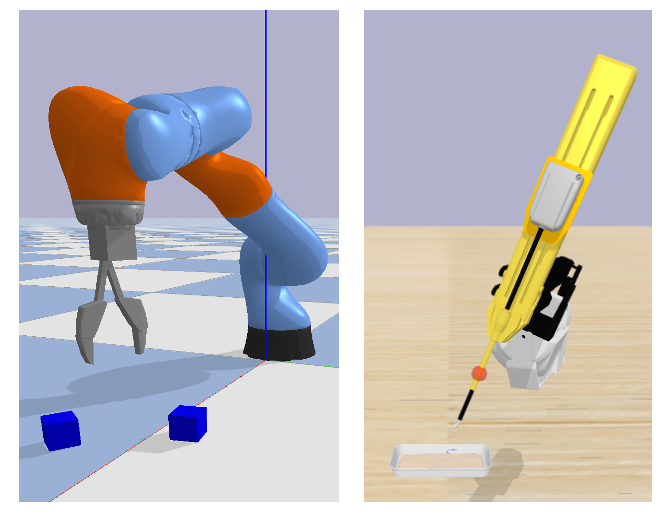}
  \caption{Rendering of simulation environments for domains of general robot manipulation (left) and surgical grasping (right).}
  \vspace{-0.45cm}
 \label{fig:rendering_sim_env}
\end{figure}

%----------------------
We consider a set of robot manipulation tasks, where a robotic arm with a gripper interacts with $M$ target objects.
The POMDP's observation includes point cloud segments of the gripper and the target objects, as well as the joint position of the gripper's jaw.
The observation is denoted as $o_t = (\{o_t^{c_i}\}^M_{i=1},o_t^g,o^\zeta_t)$, where $\{o_t^{c_i}\}^M_{i=1}$ corresponds to point cloud segments for $M$ target object, $o^g_t$ corresponds to the point cloud segment of the gripper, and $o^\zeta_t$ corresponds to the joint position of the gripper's jaws.
Each point cloud segment contains a set of points represented by 3D Cartesian coordinates.
The reference frame used in this paper is a gripper-centric frame, where its origin's position aligns with the centroid of the gripper's point clouds, and its origin's orientation aligns with the orientation of the world frame.
The POMDP's states include poses of object frames for the gripper and target objects, as well as the joint position of the jaw.
The object pose refers to the transformation that maps the reference frame of point clouds to a coordinate frame fixed to the object’s rigid body.
The state of the gripper's jaw is the same as its observation. 
We denote the POMDP's state as $s_t = (\{s_t^{c_i}\}^M_{i=1},s_t^g,s^\zeta_t)$, where $\{s_t^{c_i}\}^M_{i=1}$ and $s_t^g, s^\zeta_t$ correspond to the states of $M$ target object, the gripper and the jaw position, respectively. 
A normalized action $a_t = (a^\zeta_t, a^{xyz}_t,a^{\theta}_t)$, whose elements are in the range of $[-1,1]$, controls the actuation of the robotic gripper.
In particular, $a^{xyz}_t\in\mathbb{R}^3$ controls the Cartesian translation of the gripper along the $x$, $y$, and $z$ axes of the reference frame. 
$a^{\theta}_t\in \mathbb{R}$ controls the Cartesian orientation of the gripper along the $z$-axis of the reference frame.
Both $a^{xyz}_t$ and $a^{\theta}_t$ are delta action commands, constrained by the maximum translational and rotational magnitudes, $\Delta^{xyz}$ and $\Delta^{\theta}$, respectively. 
$a^\lambda_t\in\mathbb{R}$ directly maps to the jaw position.
Sparse delayed rewards are defined in the POMDP of robot manipulation.
The reward function is a binary function $R(s_t,a_t)\in \mathbb{1}$ where the reward is $1$ if the task is successful and $0$ otherwise.
The task is terminated only when $t$ reaches the maximum horizon $H_{max}$ or the task is successful.
In the following, we elaborate on two robot domains (see Fig. \ref{fig:rendering_sim_env}) investigated in our paper: general robot manipulation and surgical grasping. 

\subsubsection{General Robot Manipulation}
\label{sec: task_env_general_robot_manipulation}
The domain of general robot manipulation includes three representative tasks commonly studied in the field: (i) \texttt{Block Pull}, (ii) \texttt{Block Pick}, and (iii) \texttt{Drawer Open}.
The tasks are developed upon the BulletArm suit \cite{wang2022bulletarm} and the work in \cite{nguyen2023equivariant}. 
The target objects are cuboids for the first two tasks and drawers for the last task (see Fig.\ref{fig:target_object}).
% Task Details
Each robotic task consists of two identical target objects, with one being movable and the other fixed.
The movability, as an additional POMDP state, is unobservable for controllers and remains unchanged during a rollout.
Because the state of movability does not influence our formulation, we omit it from the formal equations.
% Goals
In the block-pull task, the robotic gripper must pull a movable cuboid until it touches a stationary one.
In the block-pick task, the agent is required to grasp a movable block using the gripper
In the drawer-open task, the gripper must open a closed, movable drawer.
The action space in the domain of general robot manipulation is continuous as $A_c\in \mathbb{R}^5$.
%
% Termination
In these tasks, Termination occurs if the task goal is achieved or the maximum timestep $50$ is reached. 
% Rewards
Sparse rewards $r_t$ are used, where the reward is always $0$ except for a value of $1$ if the task is successfully completed.

\subsubsection{Surgical Grasping}
\label{sec: task grasp any}
\begin{figure}[!tbp]
  \centering
  \includegraphics[width=0.7\hsize]{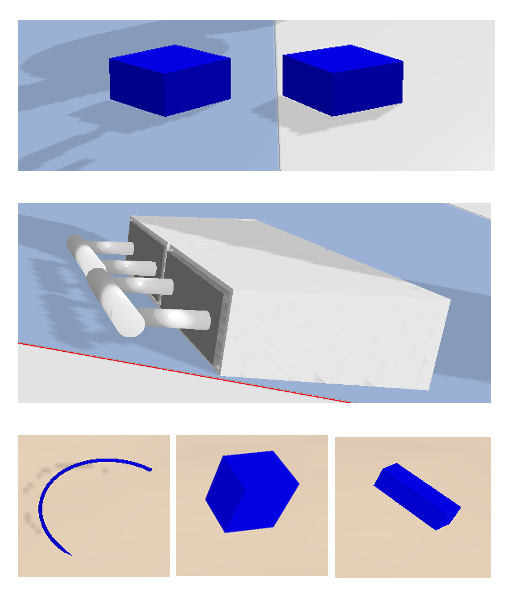}
  \caption{Target objects used in general robot manipulation and surgical grasping domains.  (\textit{Top}) A pair of identical cuboids used in \texttt{Block Pull} and \texttt{Block Pick}.  
(\textit{Middle}) A pair of identical drawers used in \texttt{Drawer Open}.  
(\textit{Bottom}) A needle, a block, and a rod used in \texttt{Surgical Grasp Any}.}
  \vspace{-0.45cm}
 \label{fig:target_object}
\end{figure}

We investigate a surgical task, \texttt{Surgical Grasp Any}, in the field of surgical autonomy, where a surgical robot is required to grasp surgical objects in the robot-assisted surgery.
The task is developed based on our previous work in \cite{lin2025gasv2}. 
The robotic arm is a 6-DOF manipulator, \textit{patient side manipulator} (PSM).
Point clouds are observed from a stereo camera, which is fixed during a rollout.
A needle, a block, and a rod are used as the target objects (see Fig. \ref{fig:target_object}). 
The occurrence probabilities of the needle, the block, and the rod are $0.5$, $0.25$, and $0.25$, respectively. 
Compared to the tasks in general robot manipulation, we discretize the agent's action.
The discrete action space, containing 9 discrete actions, is denotes as $A_{\text{d}}=\{\pm a^x_{d},\pm a^y_{d},\pm a^z_{d},\pm a^{\theta}_{d},a^{\lambda}_{d}\}$, where $a^x_{d}=[1,0,0,0,0]$, $a^y_{d}=[0,1,0,0,0]$, $ a^z_{d}=[0,0,1,0,0]$, $a^{\theta}_{d}=[0,0,0,1,0]$, and $a^{\lambda}_{d}$ correspond to the $x,y,z$ translation action, rotation action and jaw toggle action, respectively. If the jaw toggle $a^{\lambda}_{d}$ is activated, the jaw will move based on the current jaw state. If the jaws are opened, it closes the jaw with a robot command $[0,0,0,0, -1]$. Otherwise, if the jaws are closed, it opens the jaw with a robot command $[0,0,0,0, 1]$.
In the task, sparse rewards are used, and termination occurs if the task goal is achieved or the maximum timestep $80$ is reached. 
Details can be found in \cite{lin2025gasv2}.

\subsection{Task Prior of Robot Manipulation}
We leverage task priors in robot manipulation to introduce inductive biases and simplify the problem.
In the following, we elaborate on three priors we used in this paper.

\subsubsection{Simplification for Reward and Observation Functions}
\label{sec:assumption_Simplification for Reward and Observation Functions}
In the first prior, we simplify the POMDP formulation based on the problem formulation in our robot manipulation tasks. The simplifications are as follows: (i) the reward function depends only on the current state and (ii) the observation function is independent of the previous action. Therefore, we denote the simplified functions of reward and observation as $R(s_t)$ and $O(s_t, o_t)$.

\subsubsection{Invariant Transformation in Observation Function}
\label{sec:assumption_Invariant Transformation in Observation Function}
The second prior assumes that applying the same transformation to both an object's pose and its point cloud results in an invariant probability under the observation function.
In particular, let $g_to_t = (\{W^{c_i}o_t^{c_i}\}^M_{i=1},W^{g}o_t^g,o^\zeta_t)$ and $g_ts_t = (\{W^{c_i}s_t^{c_i}\}^M_{i=1},W^{g}s_t^g,\zeta_t)$, where $W^{c_i}$ and $W^{g}$ are matrices of transformation via multiplication correspond to target object $c_i$ and the gripper, respectively. 
Then, we have the invariant probability as $O(s_t,o_t) = O(g_ts_t,g_to_t)$.
This invariant prior reflects that transforming the reference frame does not alter the physical outcomes in robot manipulation tasks.

\subsubsection{Deterministic Transition Function in Robot Manipulation}
\label{sec:Deterministic Transition for Approaching}

The third prior assumes that, for certain states $s_t \in S$ encountered in robotic manipulation tasks, the transition function is deterministic and defined as:
\begin{equation}
T(s_t, a_t, s_{t+1}) = \mathbb{I}(s_{t+1} = s^*_{t+1} \mid a_t, s_t)\, Pr(a_t, s_t),
\end{equation}
where $\mathbb{I}\{\cdot\}$ is the indicator function, $Pr(a_t, s_t)$ is the joint probability of the current state and action, and $s^*_{t+1} \in S$ denotes the unique next state that deterministically follows from $s_t$ and $a_t$.
Such deterministic transitions frequently occur in robotic manipulation, especially when there is no interaction between the gripper and target objects, and the environment is free from external disturbances. 
In these scenarios, the next state can be accurately predicted: the gripper’s pose, relative to the world frame, can be inferred from the delta action applied to the end-effector; the target object's pose remains unchanged; and the gripper’s jaw state is fully determined by the jaw action.
We define the deterministic next state as a tuple:    
\begin{equation}
s^*_{t+1} = (\{s_{t+1}^{c_i*}\}_{i=1}^M,\, s_{t+1}^{g*},\, s_{t+1}^{\zeta*}), 
\end{equation}
where $s_{t+1}^{c_i*}$, $s_{t+1}^{g*}$, and $s_{t+1}^{\zeta*}$ are the predicted next states for the $i$-th target object, the gripper, and the gripper’s jaw, respectively.
Notably, instead of using the world frame, we are using our gripper-centric reference frame defined in Sec. \ref{sec:POMDP in Robot Manipulation}.
For the target object state, a translation action $a^{xyz}_t$ induces a displacement of $-a^{xyz}_t \Delta^{xyz}$ in the object's pose relative to our reference frame. 
The other action components $a^\theta_t$ and $a^\zeta_t$ do not affect the object. Thus, the target object’s next state is:
\begin{equation}
\label{equ:assumption_target_state_prediction}
    s_{t+1}^{c_i*} = \rho_l(-a^{xyz}_t \Delta^{xyz}) s_t^{c_i}.
\end{equation}
For the gripper, the rotation component $a^\theta_t$ rotates its pose by $a^\theta_t \Delta^\theta$ about the $z$-axis of our reference frame. Translation and jaw actions do not influence this state, leading to:
\begin{equation}
\label{equ:assumption_gripper_state_prediction}
    s_{t+1}^{g*} = \rho_\theta(a^\theta_t \Delta^\theta) s_t^g.
\end{equation}
Finally, the jaw action $a^\zeta_t$ directly determines the jaw's position in the next timestep, independent of other actions or the current state:
\begin{equation}
\label{equ:assumption_gripper_jaw_state_prediction}
    s_{t+1}^{\zeta*} = a^\zeta_t.
\end{equation}

\begin{figure*}[!tbp]
  \centering
  \includegraphics[width=1.0\hsize]{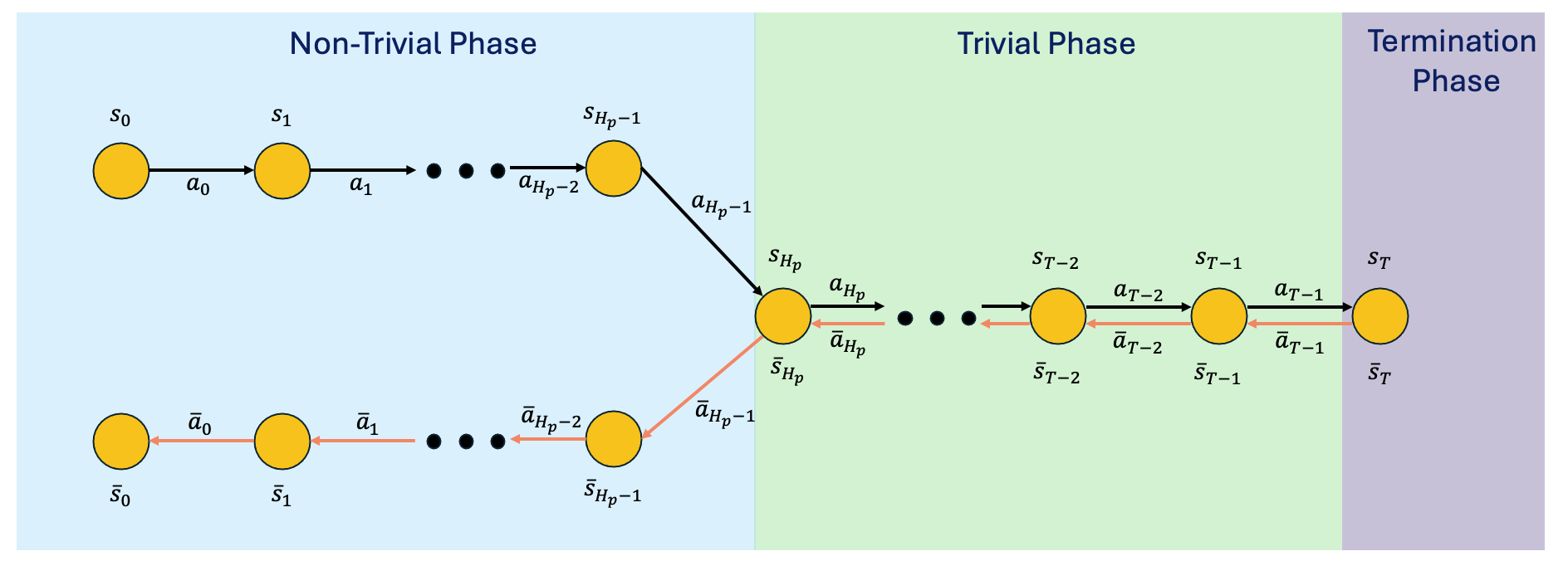}
  \caption{Graphic illustration for our multi-phase augmentation. The augmentation consists of three distinct phases: the non-trivial phase, the trivial phase, and the termination phase. Ground-truth transitions are collected in a forward time sequence as the agent interacts with the environment. In contrast, augmented transitions are generated in reverse time order, starting from the termination phase and proceeding backward through the trivial and non-trivial phases. The augmented transitions are identical to the ground-truth transitions in the trivial and termination phases, but they differ in the non-trivial phase.}
  \vspace{-0.45cm}
 \label{fig:phases_illustration}
\end{figure*}
% \begin{figure}[!tbp]
%   \centering
%   \includegraphics[width=0.6\hsize]{fig/imperfect_symmetry.png}
%   \caption{An illustration of imperfect symmetry in \texttt{Block Pick}, viewed from the top down. The orange gripper attempts to grasp a blue box-shaped target object. In the left configuration, closing the gripper results in a stable grasp. However, rotating the gripper 45 degrees around the $z$-axis may lead to an unstable grasp, potentially causing the grasp attempt to fail.}
%   \vspace{-0.45cm}
%  \label{fig:trivial_symmetry}
% \end{figure}

\section{Multi-Group Equivariant Augmentation for Robot Manipulation} 
To improve the sampling efficiency of visuomotor learning in robot manipulation, we propose a novel data augmentation, Multi-Group Equivariant Augmentation (MEA), leveraging the inductive prior of multi-group-invariant POMDP.
Effective data augmentation for visuomotor learning must satisfy three key criteria: \textit{validity}, \textit{relevance}, and \textit{diversity}~\cite{mitrano2022data}. 
Our proposed augmentation is designed to meet all three. 
Our augmentation should
comply with the rules of POMDP (validity), while also striving to cover all viable state spaces (diversity) and generalize to similar environments (relevance). 
We formulate the augmentation problem as follows: given ground-truth transitions $(a_t, r_t, o_t, \lambda_t)$, we generate augmented transitions $(\bar{a}_t, \bar{r}_t, \bar{o}_t, \bar{\lambda}_t)$.
These augmented transitions are added to the same replay buffer as the original transitions, defined as:  
\begin{equation}
    \mathcal{D} \leftarrow\{\bar{a}_t,\bar{r}_t,\bar{o}_t,\bar{\lambda}_t \}.
\end{equation}
In the following, we introduce our data augmentation from the perspectives of validity, diversity, and relevance.

\subsection{Ensuring Validity with Multi-Phase Augmentation}
\label{sec:validity}

Augmented trajectories should be feasible in the environment to satisfy validity.
Failing to meet this requirement will significantly degrade visuomotor learning performance or even result in catastrophic failures.
In the following, we define the math definition of feasibility (validity).

\noindent{\bf{Definition 4:}} 
Let a trajectory be $\{(a_t,r_t,o_t,s_t)\}_{t=0}^{T}$, we say the trajectory is \textit{feasible} if and only if $T(s_t,a_t,s_{t+1})\neq 0$, $r_t \equiv R(s_t,a_t)$, $O(a_{t-1},s_t,o_t) \neq 0$ and $b(s_0)\neq0$.

Our augmentation leverages the prior of multi-group symmetry in manipulation tasks, which generates augmented transitions with non-isometric group transform $(\bar{a}_t, \bar{r}_t, \bar{o}_t, \bar{\lambda}_t) = (g_ta_t, g_tr_t, g_to_t, g_t \lambda_t)$.
If a ground-truth trajectory is feasible, and its multi-group augmented trajectory satisfies the multi-group-invariant conditions in  (\ref{equ:condition_noniso_GPOMDP_T},\ref{equ:condition_noniso_GPOMDP_R},\ref{equ:condition_noniso_GPOMDP_O},\ref{equ:condition_noniso_GPOMDP_B}), then the augmented trajectory is also feasible, since $T(g_t s_t,g_t a_t,g_{t+1} s_{t+1}) \equiv T(s_t,a_t,s_{t+1}) \neq 0$, $g_tr_t \equiv r_t \equiv R(s_t,a_t) \equiv R(g_t s_t, g_t a_t)$, $O(g_{t-1} a_{t-1},g_{t}s_{t},g_{t}o_{t}) \equiv O(a_{t-1},s_t,o_t) \neq 0$ and $b_0(g_0 s_0) \equiv b(s_0)\neq0$.
Therefore, we need to find proper groups $g_t$ that meet the four multi-group conditions to guarantee the feasibility of the multi-group augmented trajectory.

However, in robot manipulation tasks, it is challenging to determine the feasibility of transitions in states where the robotic gripper interacts with target objects. 
For example, consider a state $s_t$ in which the robot attempts to grasp an object by closing its gripper. 
The next state $s_{t+1}$ can vary: the object may be successfully grasped, it may slip to a different position, or it may remain stationary if it is too far from the gripper's jaws.
These outcomes depend on multiple factors, such as whether the object is positioned between the gripper jaws, as well as environmental properties like surface friction and ground rigidity. 
Since these factors are typically unknown to practitioners, assessing transition feasibility becomes difficult.
Even if a ground-truth transition is feasible within such an interaction state, it remains unclear whether a transition augmented with non-trivial representations---which likely results in a similar interaction state---is also feasible.

To address this, we treat interaction states as imperfect states and only augment their transitions using the trivial representation. Since the trivial representation leaves the transition unchanged, the feasibility of the original transition is preserved. In other words, augmenting a feasible transition with the trivial representation results in a feasible augmented transition.
Our data augmentation strategy divides a trajectory rollout into three distinct phases: (i) a \textit{termination phase}, (ii) an \textit{trivial phase}, and (iii) a \textit{non-trivial phase}.
For the termination phase, we identify the transition corresponding to the final timestep of an episode ($t = T_H$). 
To distinguish between the trivial and non-trivial phases, we evaluate whether each transition satisfies the deterministic transition function assumption (see Sec.~\ref{sec:Deterministic Transition for Approaching}).
If the state $s_t$ satisfies this assumption, the transition is identified as the non-trivial phase ($t \in [0, H_p)$), where $H_p$ denotes the phase transition timestep separating the trivial and non-trivial segments.
Otherwise, the transition is identified as the trivial phase ($t \in [H_p, H)$).
In our robot manipulation tasks, the deterministic transition function holds as long as the gripper does not interact with the target objects.
We apply a gripper-height-based criterion to identify these phases in the general robot manipulation domain, and a timestep-based criterion in the surgical grasping domain.
The detailed implementation of these identification strategies is provided in the Appendix \ref{sec:appendix_Identification Between Trivial and Non-Trivial Phases}.

Another challenge in our data augmentation process is ensuring that the rewards in the augmented transitions remain identical to those in the ground-truth transitions, as required by the reward condition defined in (\ref{equ:condition_noniso_GPOMDP_R}).
Given rewards in our manipulation tasks are defined as sparse and delayed (as described in Sec. \ref{sec:POMDP in Robot Manipulation}), the augmented rewards should match this structure: a success reward of $1$ at the final timestep for successful rollouts, a failed reward $0$ at the final timestep for failed rollouts, and $0$ for all preceding timesteps in both cases.
As such, the rewards of augmented transitions at non-final timesteps are always zero. However, it is challenging to determine the success or failure rewards of augmented transitions at the final timestep, as the reward function in the POMDP is unknown.

To address the challenge of reward preserving in the setting of sparse delayed rewards, we augment the transitions in a rollout using a reversed time sequence.
In particular, we begin by augmenting the transition at the final timestep (termination phase) to ensure that the reward at this timestep remains consistent with the ground-truth trajectory.
After that, we infer the remaining augmented transitions—spanning both the trivial and non-trivial phases by progressing backward from timestep $T-1$ to $0$.
Figure \ref{fig:phases_illustration} shows the illustration for three phases.
In the following section, we elaborate on the transistions augmemted with different group representations for these three phases.
Our augmentation satisfies the multi-group conditions with proof in Appendix \ref{sec:proof_valid}.

\subsubsection{Termination Phase}
In this phase ($t=H$), transitions are augmented with the trivial representation.
We augment the state $s_t$ to be identical to that of the ground truth.
The observation is augmented with the trivial representation, denoted as $\bar{o}_t=g_to_t=\rho_0o_t= o_t$.
Given the identity augmentation for states, the augmented reward is equal to the ground truth reward  as $\bar{r}_t=r_t$. 
The augmented termination flag is also equal to the flag of ground truth as $\bar{\lambda}_t=\lambda_t$. 

%----------------------
\subsubsection{Trivial Phase} 
%----------------------
In this phase, where $t\in[H_p, H)$, the rules to augment transitions are the same as those of the termination phase, except that the action is augmented using a trivial representation, as $\bar{a}_t=g_ta_t=\rho_0a_t=a_t$.
%----------------------
\subsubsection{Non-Trivial Phase} 
In this phase, where $t\in[0, H_p)$, we augment the transitions with non-trivial representations. 
The augmented action can be represented in a tuple form as $\bar{a}_t = (\bar{a}^\lambda_t, \bar{a}^{xyz}_t,\bar{a}^{\theta}_t)$, where $\bar{a}^\lambda_t=g_ta^\lambda_t$, $ \bar{a}^{xyz}_t=g_ta^{xyz}_t$, and $\bar{a}^{\theta}_t=g_ta_t^{\theta}$.
The elements in $\bar{a}^\lambda_t, \bar{a}^{xyz}_t,\bar{a}^{\theta}_t$ can be arbitrary values within $[-1,1]$ without violating our POMDP definition. 
We use structured action to enhance the diversity and relevance of our data augmentation, as detailed in the latter Section \ref{sec:Diversity and Relevance}.

The augmented observation can be decomposed as $\bar{o}_t = (\{\bar{o}_t^{c_i}\}^M_{i=1},\bar{o}_t^g,\bar{o}^\zeta_t)$, where $\bar{o}_t^{c_i}=g_t o_t^{c_i}$ is the augmented observation of target object $i$, $\bar{o}_t^g=g_to_t^g$ is the augmented observation of the gripper, and $\bar{o}^\zeta_t =g_to^\zeta_t$ is the augmented observation of gripper state.
Similarly, the augmented observation of target object $i$ can be obtained by transforming its ground-truth observation at timestep $t$ with a translation as:
\begin{equation}
\begin{aligned}
    \bar{o}_{t}^{c_i} &=-\rho_l\big(\Delta^{xyz}\sum_{j=t}^{H_p-1}[a_j^{xyz} - \bar{a}_j^{xyz}]\big)o_t^{c_i}.
\label{equ:augment_observation_target_nontrivial}
\end{aligned}
\end{equation}
The augmented observation of the gripper can be obtained by transforming its ground-truth observation at timestep $t$ with a rotation as:

\begin{equation}
\begin{aligned}
    \bar{o}_{t}^{g} &= \rho_\theta\big(\Delta^\theta\sum_{j=t}^{H_p-1}[a_j^{\theta} - \bar{a}_j^{\theta}]\big)o_t^{g}.
\label{equ:augment_observation_gripper_perfect}
\end{aligned}
\end{equation}
The augmented observation of the gripper state remains identical with that of ground truth using the trivial representation as $\bar{o}^\zeta_t =\rho_0o^\zeta_t=o^\zeta_t$.
The derivation of $\bar{o}_{t}^{c_i}$ and $\bar{o}_{t}^{g}$ can be found in the proof in Appendix \ref{sec:proof_valid}.

We apply the same augmentation rules for reward and termination as in the termination and trivial phases, where the augmented reward and termination flag remain identical to their ground-truth counterparts.

\subsection{Enhancing Diversity and Relevance via Structured Action}
\label{sec:Diversity and Relevance}

To improve visuomotor learning, augmented trajectories that satisfy the validity constraint should be carefully designed to maximize both diversity and relevance.
In terms of diversity, it is crucial that these trajectories provide sufficiently dense coverage of the state space. 
Without adequate sampling, the learned controller may fail to generalize to novel states encountered during deployment.
To ensure relevance, the visuomotor policy trained in simulation must generalize effectively to real-world robots. 
Although the augmented action can be selected arbitrarily within the range of $[-1, 1]$, it may lead to abrupt or jerky gripper movements if the policy does not consider the continuity of actions between consecutive timesteps.
Such motions may cause excessive acceleration, potentially leading to wear or damage to the robot’s motors.

 \begin{figure}[!tbp]
  \centering
  \includegraphics[width=1.0\hsize]{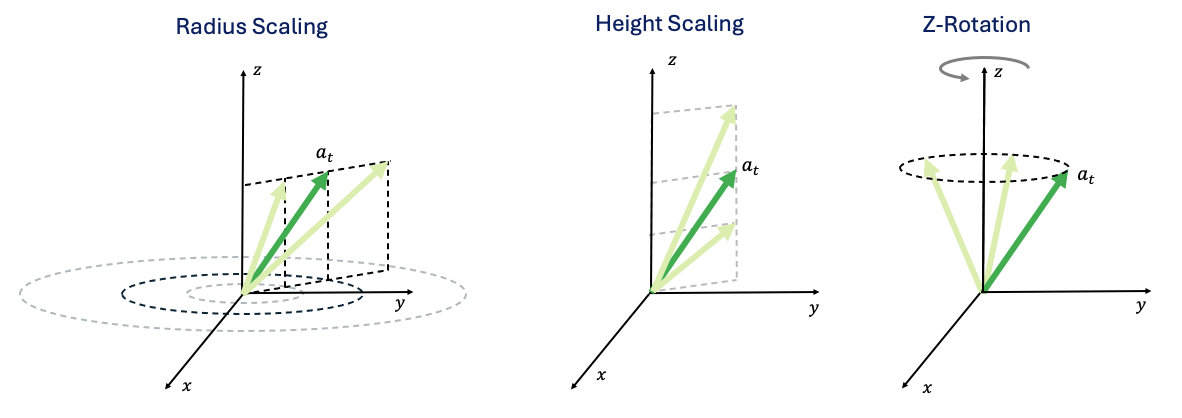}
  \caption{Radius scaling, height scaling and z-rotation for the augmented translation action. The arrows in dark green are the vectors of ground-truth translation action $a_t^{xyz}$ and the arrows in light green are the augmented vectors after applying their corresponding transformation.}
  \vspace{-0.45cm}
 \label{fig:augmented_action_illustration}
\end{figure}

\begin{figure*}[!tbp]
  \centering
  \includegraphics[width=1.0\hsize]{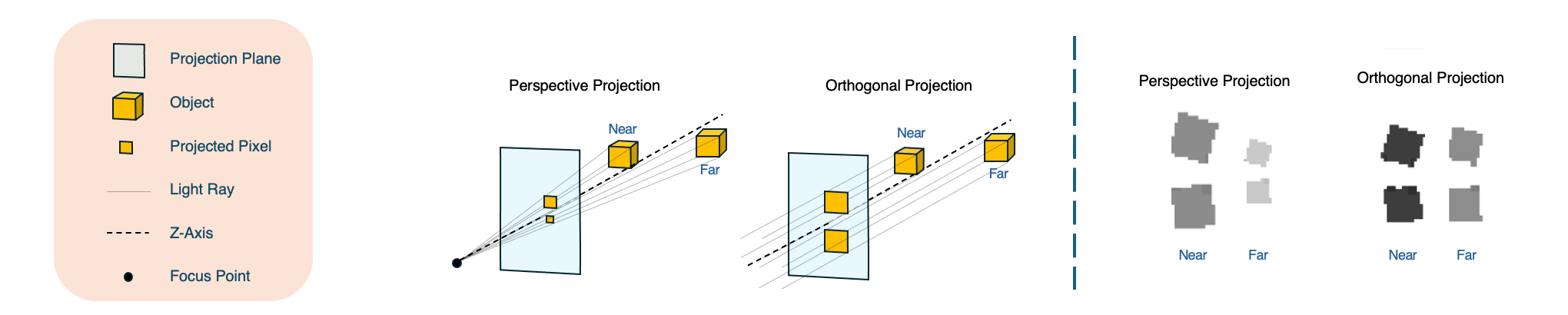}
  \caption{
 On the left, we illustrate the principles of image formation for both perspective and orthographic projections. Two camera-to-object distances---one far and one near---are compared. In perspective projection, the image is formed by radiant beams converging at a focus point, while in orthographic projection, the image is formed by parallel beams.
On the right, we show depth images $I^z$ captured at two different camera-to-object distances using both projection types.
While perspective projection does not preserve consistent shape scaling, orthographic projection maintains shape equivariance.
  }
  \vspace{-0.45cm}
 \label{fig:orthogonal_projection}
\end{figure*}

To this end, we propose structured action augmentation to enhance both the diversity and relevance of our augmentation. 
Notably, all components of the augmented transitions that meet the validity conditions (as defined in Section \ref{sec:validity}) are fully specified, except for the augmented action in the non-trivial phase.

Therefore, we augment this action in a structured manner using four controlling variables to enhance both diversity and relevance.

In particular, for augmented translation action, we first scale the ground-truth translation action $a_t^{xyz}$ along the radial and height directions. 
The radial direction corresponds to the projection of the ground-truth action onto the $x-y$ plane, while the height direction is aligned with the $z$ axis.
After scaling, we rotate the resulting vector around the $z$-axis.
Fig. \ref{fig:augmented_action_illustration} demonstrates these three types of transformation for the augmented actions.
In summary, the augmented translation action is formulated as
\begin{equation}
\begin{aligned}
     \bar{a}_t^{xyz} &= \rho_\theta(\delta_t^{\text{rot}})(k^{xyz}_t\odot a_t^{xyz}),
\end{aligned}
\end{equation}
where $\odot$ is the element-wise multiplication operator;
$\delta_t^{rot}\in[0, 2\pi)$ is a scalar variable controlling the rotation of augmented translation action at time $t$; $K^{xyz}_t$ is a three-element vector at time $t$ as 
\begin{equation}
    k^{xyz}_t = [\frac{a_t^{x}}{\sqrt{a_t^{x^2}+a_t^{y^2}}}\delta_t^r,\frac{a_t^{y}}{\sqrt{a_t^{x^2}+a_t^{y^2}}}\delta_t^r,\delta_t^z] ^T,
\end{equation}
where the first two elements of $k^{xyz}_t$ scale the ground-truth action in a radius direction and the last element scales the action in the $z$-axis direction; $\delta_t^r\in \mathbb{R}$ and $\delta_t^z\in \mathbb{R}$ are two variables controlling scaling the radius and $z$-axis element, respectively; $a_t^x$, $a_t^y$ and $a_t^y$ are elements in the ground-truth action $a_t^{xyz}$, corresponding to $x$, $y$ and $z$ translation, respectively.

The augmentation of rotational action $\bar{a}_t^{\theta}$ is obtained by adding noise to the ground-truth rotation action $a_t^{\theta}$ as
\begin{equation}
    \bar{a}_t^{\theta} = a_t^{\theta} + \delta_t^{\theta}\epsilon,
\end{equation}
where $\epsilon$ is a random noise sampled from a uniform distribution with range of $[-1,1]$ and $\delta_\theta$ is a variable controlling the scaling of noise in augmented rotation action. 
The elements of both $\bar{a}_t^{xyz}$ and $\bar{a}_t^{\theta}$ are clipped to the range of $[-1,1]$ for action normalization.

After structurally defining the augmented action in the non-trivial phase, the augmented transitions are fully determined by four structured variables $\delta_t^r$, $\delta_t^z$, $\delta_t^\text{rot}$, and $\delta_t^\theta$.
These variables are mutually independent across all timesteps.
Therefore, we define a group as $g_t = (\delta_t^r$, $\delta_t^z$, $\delta_t^{\text{rot}}$, $\delta_t^\theta)\in G_t$ for timestep $t$, where $G_t \subset R^{4}$ is subspace of a 4-dimension real space.
Since  the group-augmented trajectory satisfies the multi-group-invariant conditions in  (\ref{equ:condition_noniso_GPOMDP_T},\ref{equ:condition_noniso_GPOMDP_R},\ref{equ:condition_noniso_GPOMDP_O},\ref{equ:condition_noniso_GPOMDP_B}) for all $g_t\in G_t, t\in [0,H] $,
the POMDP is multi-group-invariant with respect to $\boldsymbol{G_H}=\{G_t\}_{i=0}^H$. 

The two robot domains investigated in this paper have different types of action spaces---continuous for general robot manipulation and discrete for surgical grasping.
Thus, we design different group spaces $\boldsymbol{G_H}$ for these domains.
For general robot manipulation, both $\delta_t^r$ and $\delta_t^z$ are uniformly sampled from the range of $[0.75, 1]$; 
$\delta_t^{\text{rot}}$ is uniformly sampled from a rotation range of $[0, 2\pi]$;
$\delta_t^\theta$ is sampled uniformly from a range of $[0, 0.3]$.
Thus, the group element is a four-element tuple $g_t = (g^1_t,g^2_t,g^3_t,g^4_t)$ and the group space is defined as $G_t=\{g_t \in G_t: 0.75<g^1_t,g^2_t<1,\ 0<g^3_t<0.3,\ 0<g^4_t<2\pi \}$ for timestep $t$.

For surgical grasping,  if the ground-truth action is within a set $A'_{\text{d}}=\{\pm a^z_{d},\pm a^{\theta}_{d},a^{\lambda}_{d}\} \subset A_{\text{d}}$, then the augmented action is a trivial action $\bar{a}_t=\rho_0a_t=a_t$.
If the ground-truth action is not within $A'_{\text{d}}$, we generate a trivial action $\bar{a}_t=\rho_0a_t=a_t$ with a probability of $0.7$, while generating a rotated action with a probability of $0.3$.
The rotated action is obtained by setting $\delta_t^{\text{rot}}$ uniformly sampled from a set $\mathrm{C}_4 = \{0,\frac{\pi}{2},\pi,\frac{3\pi}{2} \}$. Notably, the rest variables will not transform the action by setting $\delta_t^r=1,\delta_t^z=1$ and $\delta_t^\theta=0$. Thus, the group element is a real number and the group space is defined as $G_t=\{g_t : g_t \in \mathrm{C}_4\}$ for timestep $t$.
Fig. \ref{fig:mea_space} shows the sampling space of our defined $\boldsymbol{G_H}$ in a demonstration trajectory.

\section{Integrating Multi-Group Equivariant Augmentation With Off-Policy RL Methods}
\label{sec: Integrating Multi-Group Equivariant Augmentation With Off-Policy RL Methods}
In general, our augmentation strategy can be applied to any off-policy RL method to improve sampling efficiency by augmenting each ground-truth trajectory stored in the replay buffer.
We demonstrate this integration using two representative off-policy approaches: a model-free RL method and a model-based RL method. These are applied to the two domains investigated in this study.
By simply augmenting the demonstration trajectories used by these methods, we observe consistent improvements in sampling efficiency.

For the domain of general robot manipulation, we integrate a state-of-the-art (SOTA) model-free RL method, \texttt{Equi-RSAC}, in the benchmark.
Equi-RSAC is deployed in tasks of general robot manipulation leveraging isometric symmetry as an inductive prior \cite{nguyen2023equivariant}.
In particular, POMDPs in the manipulation tasks are group-equivariant with respect to an isometric group $G=\mathrm{SO}(2)$.
Meanwhile, the POMDP is also multi-group-invariant with respect to $\boldsymbol{G_H}=\{G_t\}_{i=0}^H$ as shown in Sec. \ref{sec:Diversity and Relevance}. 
Notably, the isometric group $G$ is independent of the group set $\boldsymbol{G_H}$. 
This is because $G$ applies a global isometric transformation to all task-related objects, which can be viewed as transforming the reference frame.
Consequently, an augmented trajectory with non-isometric transformation of $\boldsymbol{G_H}$ can be arbitrarily rotated around the $z$-axis of the reference frame without affecting the physical outcomes.
Therefore, we can augment the group set $\boldsymbol{G_H}$ to a new group set $\boldsymbol{G_H'}=\{G'_t = G_t \times G\}_{i=0}^H$.
The POMDPs are multi-group invariant to the new group set $G'_t$.
As such, we deploy Equi-RSAC, an equivariant version of RSAC \cite{ni2021recurrent}, to ensure that the actor network is equivariant to the isometric group $\mathrm{C}_4$—a subgroup of $\mathrm{SO}(2)$—while the critic network remains invariant with respect to $\mathrm{C}_4$.
The isometric inductive prior defined by $\mathrm{C}_4$ is directly encoded at the neural network level.
In addition to Equi-RSAC, both observations and actions in each episode are augmented with a random isometric rotation in $\mathrm{SO}(2)$, following the approach in \cite{nguyen2023equivariant}. 
For each transition stored in the replay buffer, four rotational augmentations are applied. 
To exploit the multi-group symmetry prior, we generate six augmented trajectories using our proposed MEA for each demonstration rollout. 
Importantly, all augmented episodes from MEA are applied with the same isometric $\mathrm{SO}(2)$ augmentation.
This integration of MEA with the equivariant model and isometric augmentation enables the effective incorporation of the multi-group invariance prior into the learning process.

For the surgical grasping domain, we leverage a SOTA model-based RL baseline, \texttt{GASv2} in \cite{lin2025gasv2}.
GASv2 adopts a model-based RL framework built on DreamerV2 \cite{hafner2020mastering}.
To enhance performance in surgical grasping, GASv2 leverages a hybrid control architecture that integrates traditional controls---such as PID control---with a learned visuomotor policy.
Additionally, domain randomization is applied to GASv2 to improve robustness against environmental disturbances.
In this work, we integrate GASv2 with our data augmentation, MEA. 
This integration enables the policy to exploit the inherent multi-group symmetries present in the surgical grasping task.
We augment 200 trajectories using our MEA method for each demonstration rollout. 
In the surgical grasping domain, we do not apply the isometric group $G$. 
This is due to the point clouds being captured from a camera with a non-top-down pose, 
making it unclear how to effectively incorporate isometric transformations such as those in $\mathrm{SO}(2)$ within this setting.
We leave the exploration of isometric group transformations in this domain to future work.

%----------------------
\section{ Voxel-Based Orthogonal Projection for Equivariant Image Representation}
%----------------------
Previous visuomotor learning methods typically use images with perspective projection as the controller's visual input.
For example, RGB images from stereo cameras follow the perspective projection model, where image pixels are generated by radial rays that converge toward the camera's focal point.
Fig. \ref{fig:orthogonal_projection} illustrates the principle of perspective projection.
However, images with perspective projection fail to maintain \textit{equivariance} for the image encoder based on convolutional neural networks (CNNs).
Equivariance refers to the property where translating an object in the workspace and then encoding the resulting image should yield the same feature maps as encoding the original image and then translating the feature maps.
Convolutional neural networks are inherently equivariant to translations along the $x$ and $y$ axes, where object movement results in corresponding pixel shifts in the image.
However, they are not equivariant to translations along the $z$ axis, which scale the object’s size in the image, breaking equivariance \cite{cohen2016group}.
Fig. \ref{fig:orthogonal_projection} demonstrates the scaling effects of depth images as the object translates in the $z$ axis of the camera's frame.

We propose a novel image orthogonal projection for image representation in visuomotor learning.
Our projection pipeline includes two steps: voxelization and orthogonal projection.

First, we evenly rasterize the 3D Cartesian workspace into $N_{voxel}\times N_{voxel}\times N_{voxel}$ identical cuboids along $x$, $y$, and $z$ axes, where $N_{voxel}\in \mathbb{R}^+$ is the number of voxel along an axis\cite{xu2021voxel}.
This process is known as voxelization.
These cuboids, called voxels, are indexed as $B=\{b_{x,y,z}\}$, where $b_{x,y,z}$ represent a voxel with index coordinates $(x,y,z)$ along $x,y$ and $z$ axes. 
Afterward, we identify the voxels that contain at least one point from the point clouds in $o_t$.
The identified voxels are defined as $B'=\{b_{x,y,z}: b_{x,y,z} \text{ contains points in } o_t \}\subseteq B$.

After voxelization, we orthogonally project the identified voxels $B'$ along the $z$ axis of the voxels.
The projected light rays are parallel with the $z$-axis as shown in Fig. \ref{fig:orthogonal_projection}.
Let $B_{\bar{x},\bar{y}} = \{ b_{x,y,z}\in B': x=\bar{x}, y=\bar{y}\}$ be a subset of $B'$ contains the identified voxels at the coordinates of $\bar{x}$ and $\bar{y}$.
We obtain a projected depth image $I^{proj}=\{d_{\bar{x},\bar{y}}\}^{N_{voxel}\times N_{voxel}}$, where each pixel $d_{\bar{x},\bar{y}}$ with image coordinate $(\bar{x},\bar{y})$ is given by 
\begin{equation}
    d_{\bar{x},\bar{y}} = \begin{cases} \min\limits_{b_{x,y,z} \in B'_{\bar{x},\bar{y}}} z &   \text{if }B'_{\bar{x},\bar{y}} \neq \emptyset \\
    0 & \text{otherwise}
    \end{cases}.
\end{equation}
The same image representation is applied to the augmentation transitions based on the observation $\bar{o}_t$. 
The resultant projected image $I^{proj}$ is used as the visual input for the visuomotor policies.

%----------------------
\section{Experiments}

We conduct extensive experiments to evaluate the effectiveness and generality of our approach.
Our investigation spans two robotic domains: general robot manipulation and surgical grasping.
% %
In the following, we elaborate on the experiment setups and experiment results in simulation and a real robot.

\begin{figure}[!t]
\centering
\subfloat[\texttt{Block Pull}]{\includegraphics[width=1\hsize]{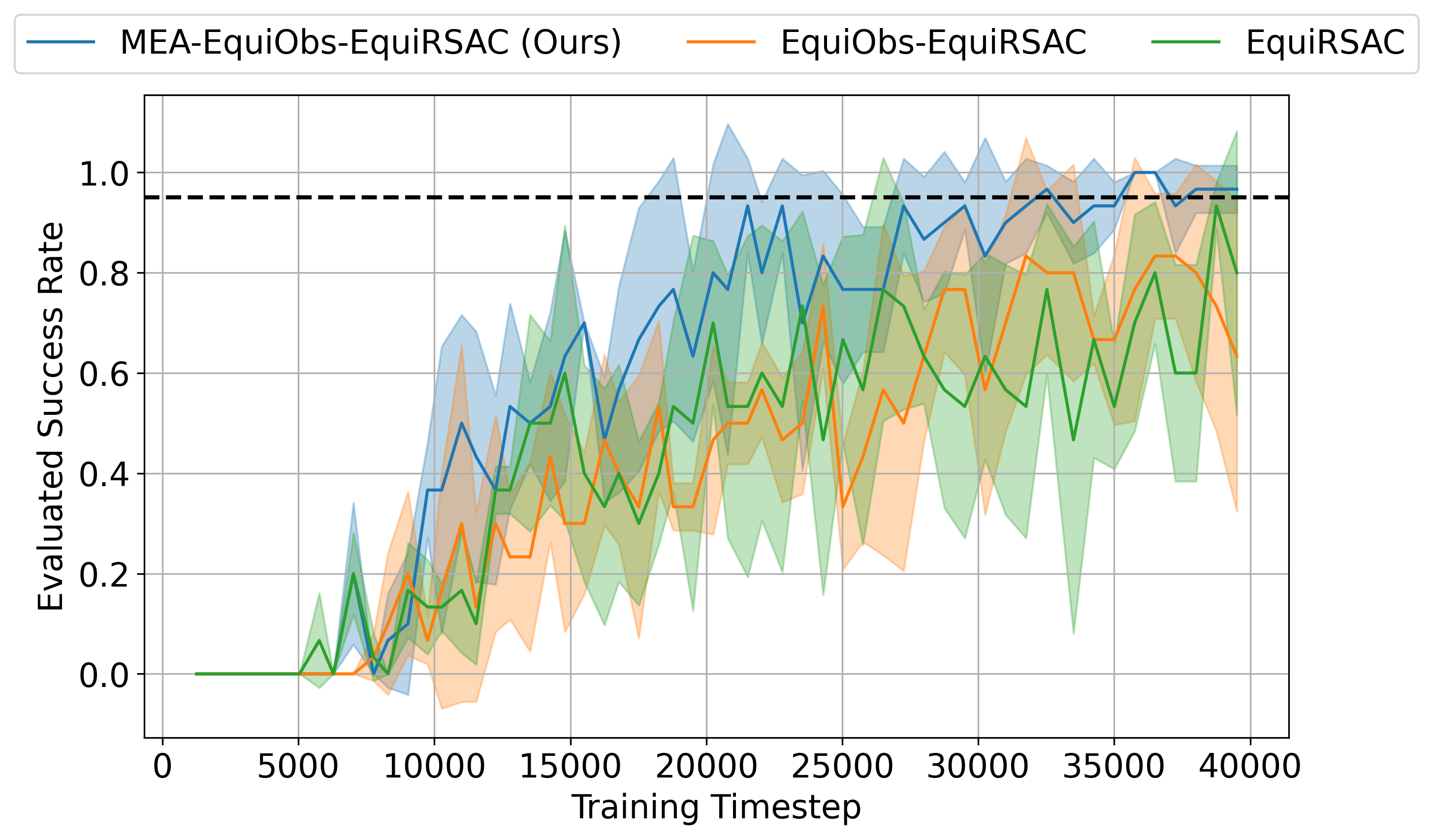}%
\label{fig:block_pull_eval}}
\hfil
\subfloat[\texttt{Block Pick}]{\includegraphics[width=1\hsize]{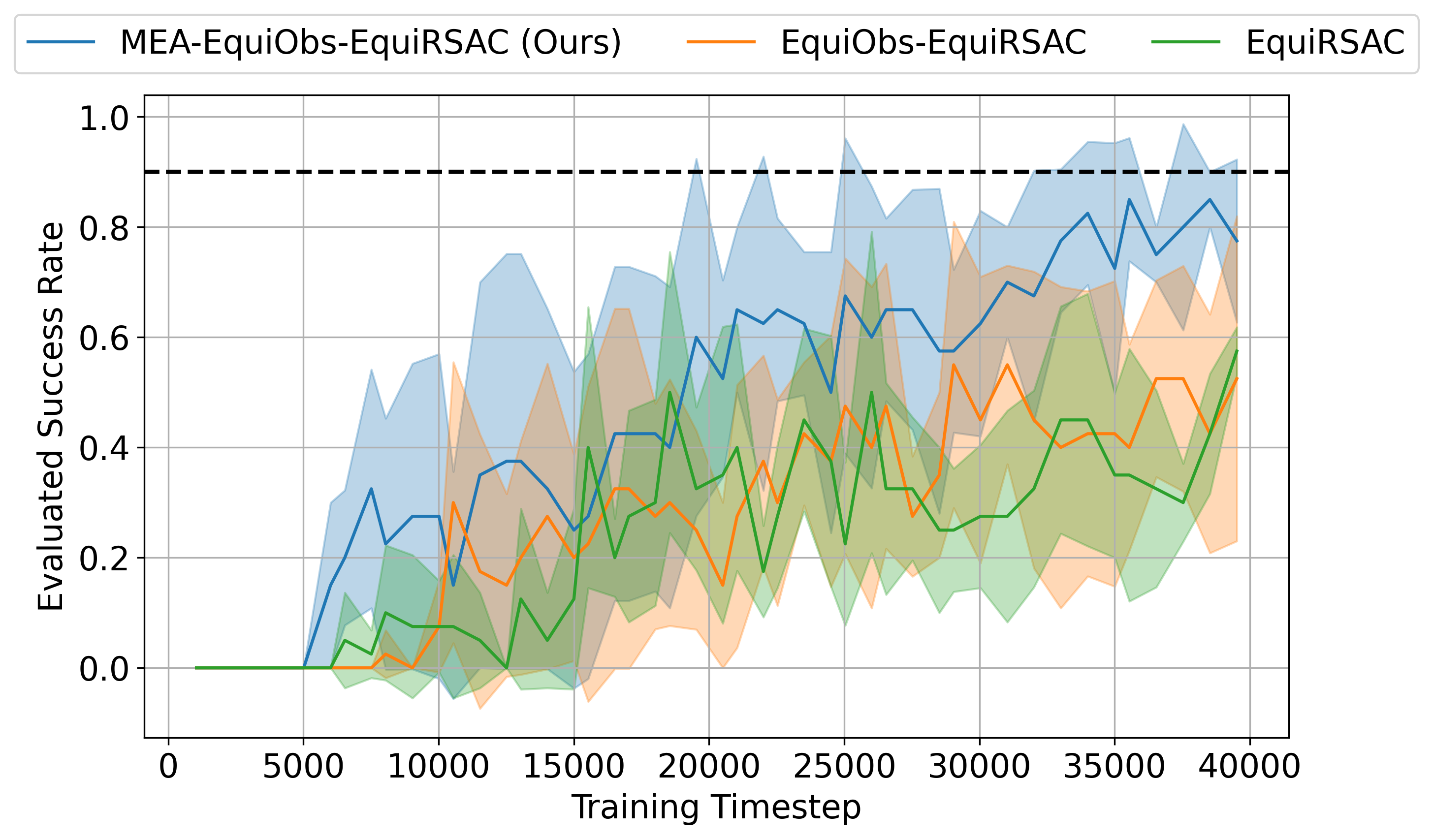}%
\label{fig:block_pick_eval}}
\hfil
% \subfloat[\texttt{Block Push}]{\includegraphics[width=0.5\hsize]{fig/block_push_eval.png}%
% \label{fig:block_push_eval}}
% \hfil
\subfloat[\texttt{Drawer Open}]{\includegraphics[width=1\hsize]{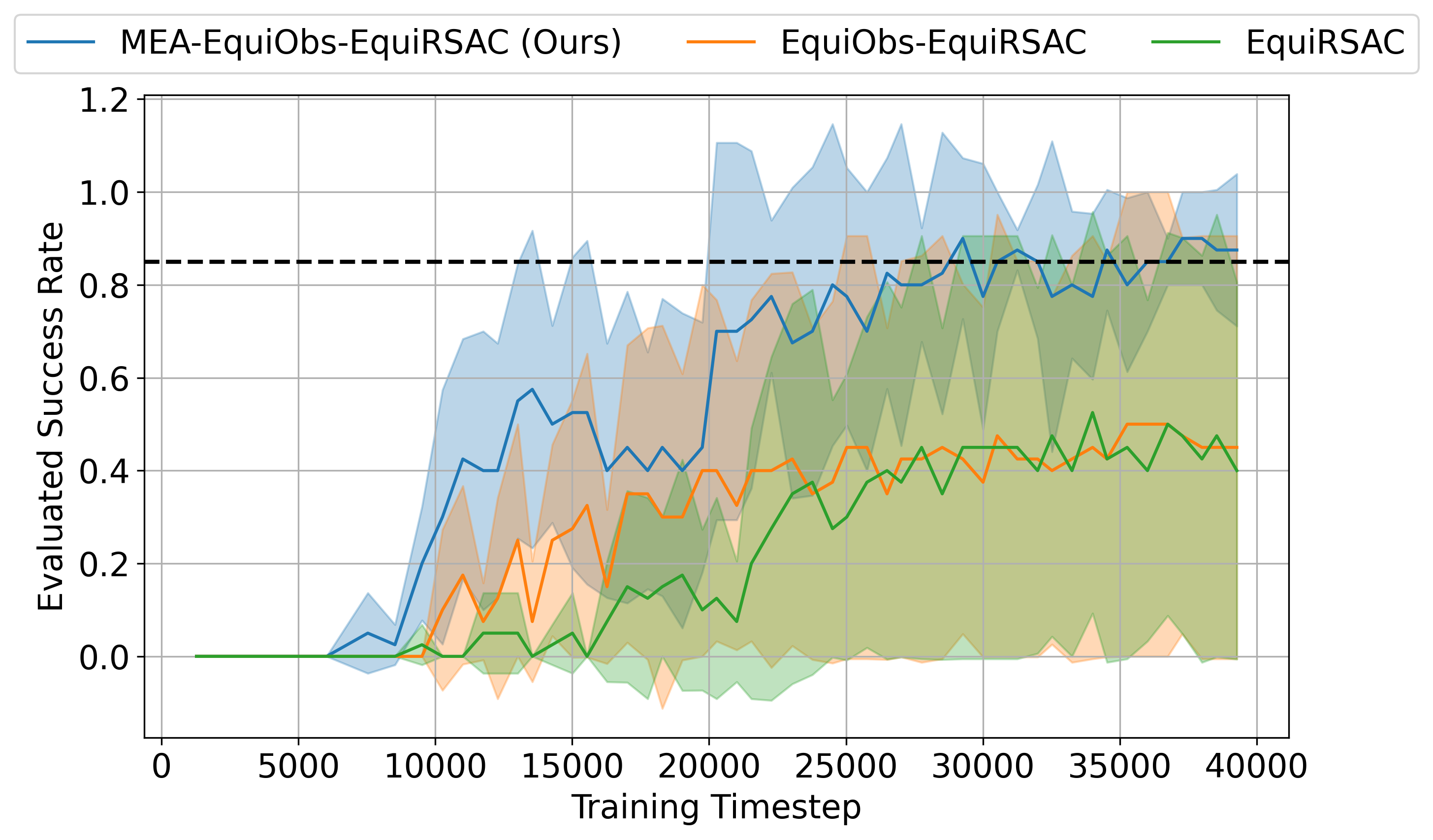}%
\label{fig:drawer_open_eval}}
\caption{Evaluated success rate in simulation experiments for 2-demo general robot manipulation.
        All baselines are trained using $2$ demonstration episodes. Each experiment is evaluated with four random seeds. Solid lines indicate the mean evaluated return, and shaded regions represent one standard deviation across seeds. The black dashed lines show the mean return of the best-performing baseline reported in~\cite{nguyen2023equivariant}, trained with $80$ demonstration episodes.}
\label{fig:general_robot_manipulation_eval}
\end{figure}
%----------------------

\subsection{Experiment Setup in Simulation}
\subsubsection{General Robot Manipulation}
We evaluate baselines in the tasks, \texttt{Block Pull}, \texttt{Block Pick}, and \texttt{Drawer Open}, introduced in \ref{sec: task_env_general_robot_manipulation}.
% demonstration
During training, fewer successful demonstration rollouts ($2$ episodes) are added for each task, compared to $80$ demonstration rollouts in the original setting in \cite{nguyen2023equivariant}.
We term our benchmark as \texttt{2-demo general robot manipulation}, and the original benchmark as \texttt{80-demo general robot manipulation}.
In one rollout, the gripper interacts with the immovable object to obtain movability information. In the other rollout, the gripper directly manipulates the movable object without interacting with the immovable object.
We train $40$K timesteps for the block-pull, block-pick, and drawer-open tasks.
% and $80$k for \texttt{Block Push}.
%
During training, the performance of baselines is evaluated using the evaluated return, obtained by averaging the returns over $10$ evaluation rollouts.
We compare our approach with the SOTA baseline, \texttt{Equi-RSAC}, introduced in Section~\ref{sec: Integrating Multi-Group Equivariant Augmentation With Off-Policy RL Methods}.
Our method extends this baseline in two key ways: by incorporating our augmentation MEA and by introducing our voxel-based equivariant image representation. 
We refer to the resulting method as \texttt{MEA-EquiObs-EquiRSAC}.
To evaluate the effectiveness of our image representation, we also evaluate an ablation baseline that enhances Equi-RSAC with our voxel-based representation but excludes data augmentation.
We refer to the ablation baseline as \texttt{EquiObs-EquiRSAC}.

\subsubsection{Surgical Grasping}
\label{sec:sim_surgical_grasp}
In the domain of surgical grasping, we evaluate baselines in the task, \texttt{Surgical Grasp Any}, in Sec. \ref{sec: task grasp any}.
In the training process, baselines are trained with $420$K timesteps; 
For each baseline, $20$ evaluation rollouts are carried out every $10$K training timestep.
We pre-fill the replay buffer with $10$ episodes of demonstration rollouts generated by a scripted policy, in contrast to approximately $1$K episodes used in the original setting~\cite{lin2025gasv2}.
We term our benchmark as \texttt{10-demo surgical grasping} and the original benchmark as \texttt{1K-demo surgical grasping}.
\texttt{GASv2} is used as the SOTA baseline in this benchmark. Our approach improves upon GASv2 by integrating it with our multi-group equivariant augmentation (MEA), resulting in the method we call \texttt{MEA-GASv2}.
Notably, GASv2, as a concurrent work to this paper, also employs orthographic projection, consistent with the image representation used in our approach.
We also compare a new baseline, \texttt{GASv2-1K}, where GASv2 is trained with $1$K episodes of demonstration rollouts in the original benchmark instead of $10$ episodes in our benchmark.

\subsection{Convergence Performance}
We compare the converged success rate in the simulation evaluation for general robot manipulation (see Fig.~\ref{fig:general_robot_manipulation_eval}).
Across all three tasks in 2-demo general robot manipulation, our approach achieves mean success rates (blue lines)---averaged over random seeds---that match the SOTA performance (dash lines) reported in the original benchmark, 80-demo general robot manipulation. 
In 2-demo general robot manipulation, compared to the SOTA baseline EquiRSAC (green lines), our method achieves mean success rate improvements of approximately $25\%$, $45\%$, and $40\%$ on the block-pull, block-pick, and drawer-open tasks, respectively.

\begin{figure}[!tbp]
  \centering
  \includegraphics[width=1.0\hsize]{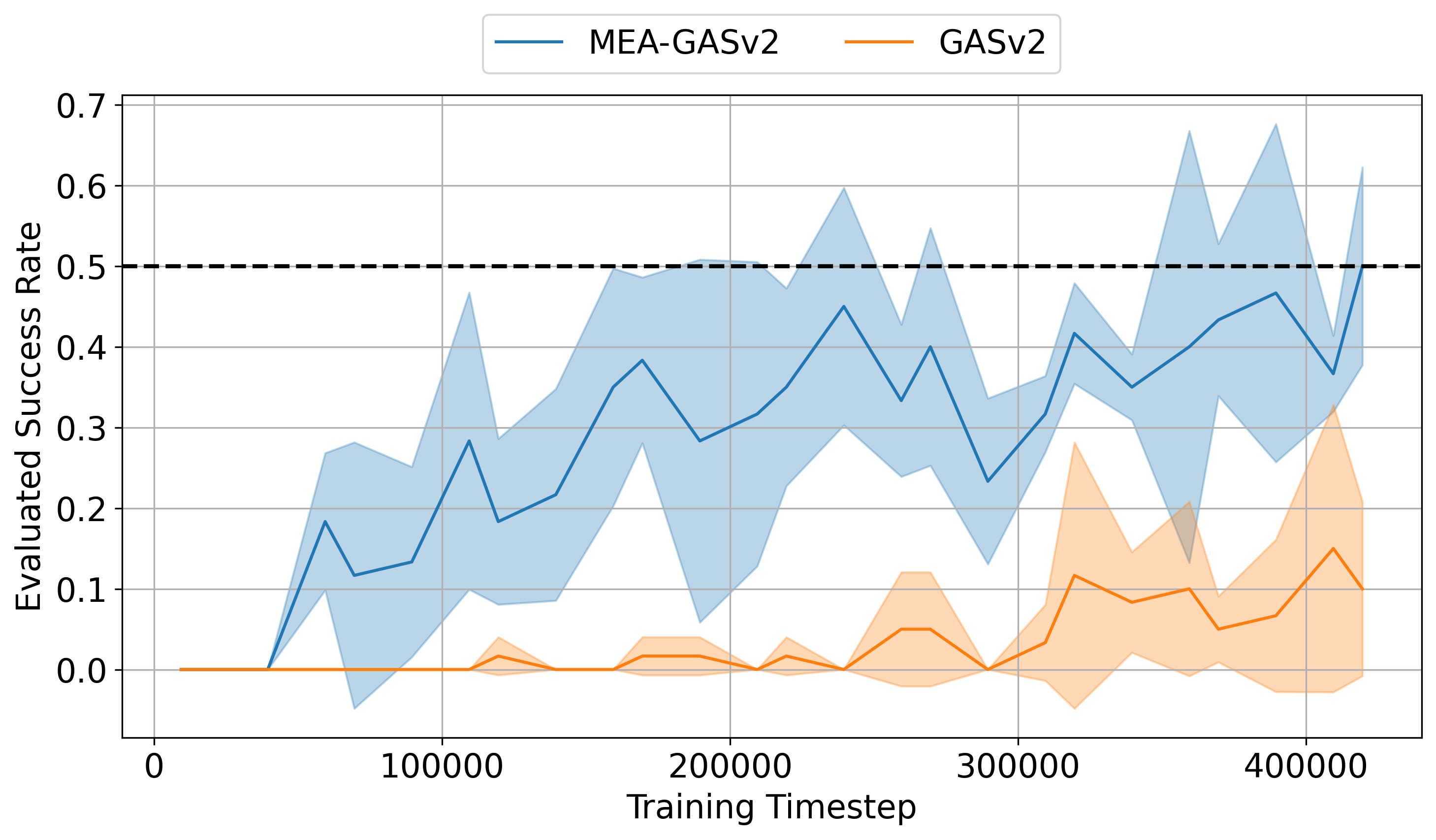}
  \caption{Evaluated success rate in simulation experiments for 10-demo surgical grasping.
        All baselines are trained using $10$ demonstration episodes. Each experiment is evaluated with three random seeds. Solid lines indicate the mean evaluated return, and shaded regions represent one standard deviation across seeds. The black dashed lines show the mean return of the best-performing baseline reported in~\cite{lin2025gasv2}, trained with $1000$ demonstration episodes.}
  \vspace{-0.45cm}
 \label{fig:surgical_grasping_sim_results}
\end{figure}

The performance of surgical grasping is shown in Fig.~\ref{fig:surgical_grasping_sim_results}.
Our method achieves a mean success rate (blue line) of approximately $50\%$ across random seeds on our 10-demo surgical grasping benchmark, matching the SOTA performance (dash line) reported in the original 1K-demo surgical grasping. 
In our 10-demo surgical grasping benchmark, our method achieves a significantly higher mean success rate of approximately $50\%$, compared to the SOTA baseline GASv2's $10\%$ (orange line).
This represents a fourfold improvement over the original SOTA baseline.
The results across two robotic domains indicate that our approach effectively enhances the convergence performance of RL.

\subsection{Sampling Efficiency}
We evaluate sampling efficiency from two perspectives: benchmark comparison and method comparison.

In the benchmark comparison, we assess reductions in demonstration data and training timesteps relative to original benchmarks. For general robot manipulation, our benchmark---2-demo General Robot Manipulation---uses the same training timesteps as the original 80-demo benchmark, but reduces the number of demonstration episodes by $97.5\%$. For surgical grasping, our 10-demo surgical grasping benchmark reduces demonstration episodes by $99\%$ compared to the original 1K-demo benchmark. 
Additionally, it reduces the training timesteps from $1100\text{K}$ to $420\text{K}$ and the training time from 116 hours (4 days and 20 hours) to 48 hours (2 days), as shown in Fig. \ref{tab:surgical_grasping_real_results}.
This results in a reduction of over $58\%$ in both training time and data.

In the method comparison, we introduce a novel metric: \textit{improvement timestep}, defined as the timestep at which the mean success rate first exceeds zero and remains positive. On the 2-demo general robot manipulation benchmark, our approach reduces the improvement timestep by over $50\%$ on tasks of block-pick and drawer-open, compared to the SOTA baseline EquiSAC. Similarly, on the 10-demo surgical grasping benchmark, our method achieves an $80\%$ reduction in improvement timestep compared to GASv2.

In summary, the consistent reduction in demonstration data, training timesteps, and improvement timestep demonstrates the sampling efficiency of our approach.

\subsection{Voxel-based Equivariant Image Representation}

We evaluate the effectiveness of our voxel-based equivariant image representation in the benchmark of 2-demo general robot manipulation.
Since the ablation baseline, EquiObs-EquiRSAC, is identical to the SOTA baseline, EquiRSAC, except for the image representation, the impact of different image representations can be directly assessed by comparing their performance.
In Fig. \ref{fig:general_robot_manipulation_eval}, both baselines exhibit comparable convergence performance across tasks.
However, the improvement timestep for the ablation baseline is observed at $10\text{K}$ for both the block-pick and drawer-open tasks, compared to $12\text{K}$ and $15\text{K}$, respectively, for the SOTA baseline. 
This represents a reduction of more than $16\%$ in improvement time.
These results demonstrate that our equivariant image representation, based on orthographic projection, enhances the speed of learning improvement, outperforming the representation based on perspective projection.

 \begin{figure}[!tbp]
  \centering
  \includegraphics[width=1.0\hsize]{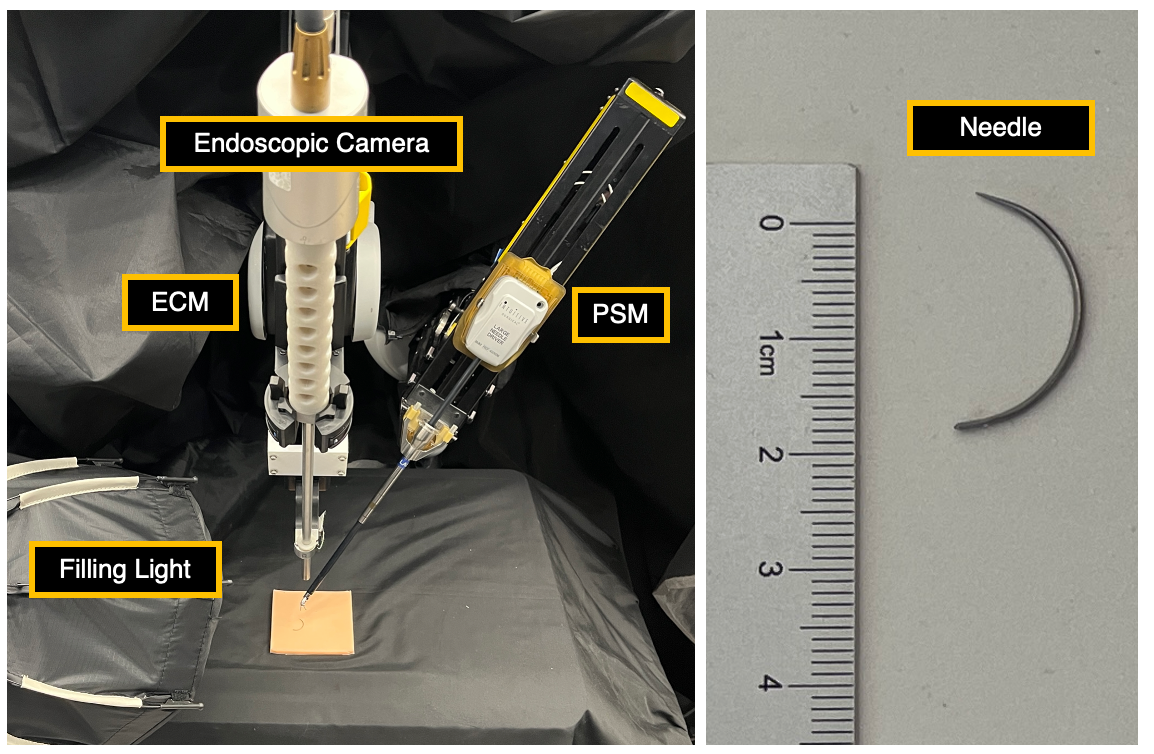}
  \caption{Surgical grasping in a real robot: setup (left) and target object (right)}
  \vspace{-0.45cm}
 \label{fig:surgical_grasp_real_setup}
\end{figure}

% \begin{table}[!tbp]
%   \centering
%   \caption{Evaluation of Surgical Grasping in Real Robot for $20$ rollouts}
%   \label{tab:surgical_grasping_real_results}
%   \begin{tabular}{lccc}
%     \toprule
%     \textbf{Baseline} & \textbf{Demonstration (Episode)} & \textbf{Success Rate (\%)} & \textbf{Score} \\
%     \midrule
%     GASv2-1K       &  1000& 70  &  0.45 \\
%     GASv2       &  10& 15  &  0.11 \\
%     GASv2-MEA   &  10& 70  &  0.47  \\
    
%     \bottomrule
%   \end{tabular}
% \end{table}

\begin{table*}[!tbp]
  \centering
  \caption{Assessment of Sim-to-Real Transfer Performance for Surgical Grasping on a Real Robot Across 20 Rollouts}
  \label{tab:surgical_grasping_real_results}
  \begin{tabular}{lcccccc}
    \toprule
    \textbf{Baseline} & \textbf{Training Timestep} & \textbf{Training Time (hour)} & \textbf{Demonstration (Episode)} & \textbf{Success Rate (\%)} & \textbf{Score} \\
    \midrule
    GASv2-1K       & 1100K  & 116   &  1000& 70  &  0.45 \\
    GASv2          & 420K  & 46  &  10  & 15  &  0.11 \\
    GASv2-MEA      & 420K  & 48 &  10  & 70  &  0.47  \\
    
    \bottomrule
  \end{tabular}
\end{table*}

\subsection{Zero-Transfer to Real Robot}

We evaluate the baseline performances on real hardware by directly transferring controllers trained in simulation to a physical robot for the surgical grasping task, without any fine-tuning. 
The experimental setup for the real-robot deployment is illustrated in Fig.~\ref{fig:surgical_grasp_real_setup}.
Controllers are deployed on a surgical robotic system consisting of a Patient Side Manipulator (PSM) for the grasping arm and an actuated Endoscopic Camera Manipulator (ECM) equipped with a stereo endoscopic camera.
The robotic arms are controlled using the open-source da Vinci Research Kit (dVRK)~\cite{kazanzides2014open}.
Stereo images are captured at 1080p resolution and center-cropped to $600 \times 600$ pixels to serve as visual observations.
Depth is estimated from stereo pairs, and point clouds are reconstructed using the camera's intrinsic parameters, following the pipeline described in~\cite{lin2025gasv2}.
A filling light is used to control illumination in the workspace.
The target object is a standard $20$\,mm surgical needle, as shown in Fig.~\ref{fig:surgical_grasp_real_setup}.
At the beginning of each rollout, the target object's pose is manually randomized within the workspace, and the gripper is initialized to a random starting pose.
After closing the gripper, it is lifted vertically by $30$\,mm.
If the gripper successfully holds the object after lifting, the finite state machine (FSM) marks the rollout as a success; otherwise, it is recorded as a failure.
Note that domain randomization is not applied during real-world deployment.
We evaluate each baseline over $20$ rollouts to compare performance.
In addition to the success rate, we also report the grasping score $s_{g}$ introduced in~\cite{lin2025gasv2}, defined as:
\begin{equation}
\label{equ:score}
     s_{g} = \begin{cases} \frac{(H - H_{max})}{H_{max}} &  \text{if the task succeeds} \\
    0 & \text{otherwise}
    \end{cases},
\end{equation}
where $H$ is the number of termination timestep and $H_{max}$ is the predefined maximum termination timestep.
The score $s_{g}$ quantifies how efficiently the controller completes the grasping task. 
A larger score value represents a faster completion.

Table~\ref{tab:surgical_grasping_real_results} presents the results from our real-robot experiment.
Our approach, GASv2-MEA, achieves a significantly higher success rate of $70\%$, compared to just $15\%$ for the SOTA baseline, GASv2.
In terms of task completion speed, as measured by the score metric, our method attains a value of $0.47$, substantially outperforming the baseline's $0.15$. 
These results demonstrate more than a threefold improvement in both success rate and performance score over the baseline.
When compared to the original benchmark baseline, GASv2-1K, our approach achieves comparable performance in terms of success rate and score.
This indicates that our method reduces the need for demonstration data without sacrificing performance in real-robot settings.

\section{Conclusion and Discussion}

In this work, we have investigated non-isometric symmetries by applying multiple independent group transformations across both spatial and temporal dimensions to relax constraints of isometric symmetry.
We introduced a novel formulation of the POMDP that integrates these non-isometric symmetry structures. 
Additionally, we proposed an effective data augmentation technique, Multi-Group Equivariance Augmentation (MEA), which was incorporated into offline reinforcement learning to improve sampling efficiency. 
Furthermore, we introduced a voxel-based visual representation that ensures translational equivariance. 
Our extensive simulations and real-robot experiments in two manipulation domains have demonstrated the significant effectiveness of our approach.

However, our approach has two primary limitations. 
First, the diversity of augmentations applied during the non-trivial phase relies on the assumption of a deterministic transition function. 
It remains an open question how to effectively extend the non-trivial phase framework to environments with stochastic transitions.
Second, to address imperfect symmetry, our augmentation method segments a ground-truth trajectory into three phases: a trivial phase, a non-trivial phase, and a termination phase.
In more complex manipulation tasks, however, a trajectory rollout could potentially be segmented into more than three phases, with multiple trivial and non-trivial phases required to capture non-isometric symmetries more effectively.
In our future work, we aim to address these two limitations and conduct systematic experiments across a broader range of robotic domains to evaluate the generality and effectiveness of our approach.
Additionally, we will explore equivariant network architectures that leverage the inductive bias of non-isometric symmetry, extending beyond our data augmentation approach.

\section*{Acknowledgments}
We especially thank Hai Nguyen for sharing valuable insights and providing guidance on the experiments.

{\appendices

\section{Proof for Multi-Group Equivariant Augmentation}
\label{sec:proof_valid}
In this section, we proof the augmented transitions using our data augmentation satisfies the multi-group conditions (\ref{equ:condition_noniso_GPOMDP_T},\ref{equ:condition_noniso_GPOMDP_R},\ref{equ:condition_noniso_GPOMDP_O},\ref{equ:condition_noniso_GPOMDP_B}).

Given the assumption of simplification for reward and observation functions in Sec. \ref{sec:assumption_Simplification for Reward and Observation Functions}, we denote the reward and observation functions in the following as $R(s)$ and $O(s_t,o_t)$, respectively. 

We begin by stating two properties of group representations. In particular, we introduce an inverse property of group representation: 
for any group representation $\rho$, the inverse of group representation w.r.t. a group element $g\in G$ can be formulated as: 
\begin{equation}
\label{equ:group inverse}
    \rho^{-1}(g)=\rho(-g). 
\end{equation}
Another property for group representation is the multiplication property: let $g_1$ and $g_2$ be two group elements and $\rho$ be the group representation, the multiplication property is given as:
\begin{equation}
\label{equ:group multiplication}
    \rho(g_1g_2) = \rho(\rho_1)\rho(\rho_2). 
\end{equation}

In the following, we proof the satisfaction of augmented transitions in the different phases of our data augmentation, including the termination phase, the trivial phase and the non-trivial phase.

For the termination phase and the trivial phase, the augmented observation, state, and action always hold the identity as $\bar{o}_t\equiv o_t$, $\bar{s}_t\equiv s_t$, and $\bar{a}_t\equiv a_t$. 
Thus, given the identity in state, observation and action, $T(s_t,a_t,s_{t+1})\equiv T(\bar{s}_t,\bar{a}_t,\bar{s}_{t+1})=T(g_ts_t,g_ta_t,g_{t+1}s_{t+1})$ satisfies (\ref{equ:condition_noniso_GPOMDP_T}), $O(s_{t},o_{t})\equiv O(\bar{s}_{t},\bar{o}_{t})=O(g_ts_{t},g_to_{t})$ satisfies (\ref{equ:condition_noniso_GPOMDP_O}), and $R(s_t)\equiv R(\bar{s}_t)=R(g_ts_t)$ satisfies (\ref{equ:condition_noniso_GPOMDP_R}).

For the non-trivial phase, the ground-truth transitions meet our assumption of the deterministic transition function in Sec. \ref{sec:Deterministic Transition for Approaching}, since the gripper does not interact with the target object in this phase. 
Therefore, given a tuple $(s_t, a_t, s_{t+1})$, the transition function $T(s_t, a_t, s_{t+1})$ as the indicator function is either $1$ or $0$.
When $T$ is one, $s_{t+1} = s^*_{t+1}$.
Let $\forall i \in[t+1, H_p]: s_{i}=s^*_{i}$, we have $T(s_i, a_i, s_{i+1}) = 1$ for all $i$.
Then, given $\forall i \in[t+1, H_p]: s_{i}=s^*_{i}$, the ground-truth state for the target object $i$ can be inferred based on (\ref{equ:assumption_target_state_prediction}) as
\begin{equation}
\label{equ:append_state_infer_ci_gt}
\begin{aligned}
     s_{H_p}^{c_i*} &= \rho_l(-a^{xyz}_{H_p-1} \Delta^{xyz}) s_{H_p-1}^{c_i}\\
     &\text{substitute } s_{H_p-1}^{c_i} = s_{H_p-1}^{c_i*}=\rho_l(-a^{xyz}_{H_p-2} \Delta^{xyz}) s_{H_p-2}^{c_i}\\
     &= \rho_l(-a^{xyz}_{H_p-1} \Delta^{xyz}) \rho_l(-a^{xyz}_{H_p-2} \Delta^{xyz})s_{H_p-2}^{c_i} \\
     & .... \\
     &= \underbrace{\rho_l(-a^{xyz}_{H_p-1} \Delta^{xyz}) \rho_l(-a^{xyz}_{H_p-2} \Delta^{xyz})..\rho_l(-a^{xyz}_{t} \Delta^{xyz})}_{H_p-t-1}s_t^{c_i}  \\
     & \text{multiplication of groups} \\
     &=\rho_l(-\Delta^{xyz}\sum_{j=t}^{H_p-1}a_j^{xyz})s_{t}^{c_i};
\end{aligned}
\end{equation}
The ground-truth state for the gripper can be inferred based on (\ref{equ:assumption_gripper_state_prediction}) as 
\begin{equation}
\label{equ:append_state_infer_g_gt}
\begin{aligned}
     s_{H_p}^{g*} &= \rho_\theta(a^{\theta}_{H_p-1} \Delta^\theta) s_{H_p-1}^{g}\\
     &\text{substitute } s_{H_p-1}^{g} = s_{H_p-1}^{g*}=\rho_\theta(a^{\theta}_{H_p-2} \Delta^\theta) s_{H_p-2}^{g}\\
     &= \rho_\theta(a^{\theta}_{H_p-1} \Delta^{xyz}) \rho_\theta(a^{\theta}_{H_p-2} \Delta^\theta)s_{H_p-2}^{g} \\
     & .... \\
     &= \underbrace{\rho_\theta(a^{\theta}_{H_p-1} \Delta^\theta) \rho_\theta(a^{\theta}_{H_p-2} \Delta^\theta)..\rho_\theta(a^{\theta}_{t} \Delta^\theta)}_{H_p-t-1}s_t^{g}  \\
     & \text{multiplication of groups} \\
     &=\rho_\theta(\Delta^\theta\sum_{j=t}^{H_p-1}a_j^{\theta})s_{t}^{g};
\end{aligned}
\end{equation}

Similarly, we can infer the augmented transitions in the same way.
Let $\forall i \in[t+1, H_p]: \bar{s}_{i}=s^*_{i}$, 
we have $T(\bar{s}_i, \bar{a}_i, \bar{s}_{i+1}) = 1$ for all $i$.
Then, given $\forall i \in[t+1, H_p]$, the augmented state for the target object $i$ is
\begin{equation}
    s_{H_p}^{c_i*}=\rho_l(-\Delta^{xyz}\sum_{j=t}^{H_p-1}\bar{a}_j^{xyz})\bar{s}_{t}^{c_i};
\label{equ:append_state_infer_ci_aug}
\end{equation}
The augmented state of gripper can be inferred as
\begin{equation}
\label{equ:append_state_infer_g_aug}
    s_{H_p}^{g*} =\rho_\theta(\Delta^\theta\sum_{j=t}^{H_p-1}\bar{a}_j^{\theta})\bar{s}_{t}^{g};
\end{equation}
We can rearrange (\ref{equ:append_state_infer_ci_aug}) and get
\begin{equation}
\label{equ:append_state_full_aug_ci}
\begin{aligned}
   \bar{s}_{t}^{c_i} &=  \rho_l^{-1}(-\Delta^{xyz}\sum_{j=t}^{H_p-1}\bar{a}_j^{xyz}) s_{H_p}^{c_i*} \\
   &\text{group inverse} \\
   & = \rho_l(\Delta^{xyz}\sum_{j=t}^{H_p-1}\bar{a}_j^{xyz}) s_{H_p}^{c_i*}  \\
   & \text{since $\bar{s}_{H_p}=s_{H_p}=s^{*}_{H_p}$, substitute (\ref{equ:append_state_infer_ci_gt}) } \\
   & = \rho_l(\Delta^{xyz}\sum_{j=t}^{H_p-1}\bar{a}_j^{xyz})\rho_l(-\Delta^{xyz}\sum_{j=t}^{H_p-1}a_j^{xyz})s_{t}^{c_i} \\
   & \text{group multiplication} \\
   & = \rho_l(-\Delta^{xyz}\sum_{j=t}^{H_p-1}a_j^{xyz}+\Delta^{xyz}\sum_{j=t}^{H_p-1}\bar{a}_j^{xyz})s_{t}^{c_i} \\
   & = -\rho_l(\Delta^{xyz}\sum_{j=t}^{H_p-1}[a_j^{xyz}-\bar{a}_j^{xyz}])s_{t}^{c_i}
\end{aligned}
\end{equation}

We can rearrange (\ref{equ:append_state_infer_g_aug}) and get
\begin{equation}
\label{equ:append_state_full_aug_g}
\begin{aligned}
   \bar{s}_{t}^{g} &=  \rho_\theta^{-1}(\Delta^\theta\sum_{j=t}^{H_p-1}\bar{a}_j^{\theta}) s_{H_p}^{g*} \\
   &\text{group inverse} \\
   & = \rho_\theta(-\Delta^\theta\sum_{j=t}^{H_p-1}\bar{a}_j^{\theta}) s_{H_p}^{g*}  \\
   & \text{since $\bar{s}_{H_p}=s_{H_p}=s^{*}_{H_p}$, substitute (\ref{equ:append_state_infer_g_gt}) } \\
   & = \rho_\theta(-\Delta^\theta\sum_{j=t}^{H_p-1}\bar{a}_j^{\theta})\rho_\theta(\Delta^\theta\sum_{j=t}^{H_p-1}a_j^{\theta})s_{t}^{g} \\
   & \text{group multiplication} \\
   & = \rho_\theta(\Delta^\theta\sum_{j=t}^{H_p-1}a_j^{\theta}-\Delta^\theta\sum_{j=t}^{H_p-1}\bar{a}_j^{\theta})s_{t}^{g} \\
   & = \rho_\theta(\Delta^\theta\sum_{j=t}^{H_p-1}[a_j^{\theta}-\bar{a}_j^{\theta}])s_{t}^{g}
\end{aligned}
\end{equation}
Since $\forall i \in[t+1, H_p]: s_{i}=s^*_{i}$ and $\bar{s}_{i}=s^*_{i}$, we have $T(\bar{s}_i, \bar{a}_i, \bar{s}_{i+1}) = T(g_is_i, g_ia_i, g_{i+1}s_{i+1}) = T(s_i, a_i, s_{i+1}) =   1$ for all $i$. 
This satisfies the multi-group transition condition in (\ref{equ:condition_noniso_GPOMDP_T}).

We augment the observation based on the assumption of invariant transformation in the observation function (see Sec. \ref{sec:assumption_Invariant Transformation in Observation Function}).
Let $W^{c_i} = \rho_l(\Delta^{xyz}\sum_{j=t}^{H_p-1}[a_j^{xyz}-\bar{a}_j^{xyz}])$, then (\ref{equ:append_state_full_aug_ci}) can be represented as $\bar{s}_t^{c_i} =W^{c_i}s_t^{c_i} $. The augmented observation for target object $i$ can be represented as
\begin{equation}
\begin{aligned}
    \bar{o}_t^{c_i} &=W^{c_i}o_t^{c_i} \\
    & =-\rho_l(\Delta^{xyz}\sum_{j=t}^{H_p-1}[a_j^{xyz}-\bar{a}_j^{xyz}]) o_t^{c_i},
\end{aligned}
\end{equation}
which gets the same equation in (\ref{equ:augment_observation_target_nontrivial}).

Similarly, let $W^{g}=\rho_\theta(\Delta^\theta\sum_{j=t}^{H_p-1}[a_j^{\theta}-\bar{a}_j^{\theta}])$, then (\ref{equ:append_state_full_aug_g}) can be represented as $\bar{s}_t^{g} =W^{g}s_t^{g} $.
The augmented observation for the gripper can be represented as
\begin{equation}
\begin{aligned}
    \bar{o}_t^{g} &=W^{g}o_t^{g} \\
    & =\rho_\theta(\Delta^\theta\sum_{j=t}^{H_p-1}[a_j^{\theta}-\bar{a}_j^{\theta}]) o_t^{g},
\end{aligned}
\end{equation}
which gets the same equation in (\ref{equ:augment_observation_gripper_perfect}).

Since $\bar{o}_t = g_to_t = (\{W^{c_i}o_t^{c_i}\}^M_{i=1},W^{g}o_t^g,o^\zeta_t)$ and $\bar{s}_t = g_ts_t = (\{W^{c_i}s_t^{c_i}\}^M_{i=1},W^{g}s_t^g,\zeta_t)$, then we have the invariant probability as $O(s_t,o_t) = O(g_ts_t,g_to_t)$ based on the assumption of invariant transformation in observation function.
This satisfies the multi-group observation condition in (\ref{equ:condition_noniso_GPOMDP_O}).

For reward, since the task is not successful in this state, $R(s_t)=R(\bar{s}_t)=R(g_ts_t)=0$, which satisfies the multi-group reward condition (\ref{equ:condition_noniso_GPOMDP_R}).

For the initial state, since the states of both gripper and target objects are randomly initialized, $b(s_0)=b(\bar{s}_0)=b(g_0s_0)$, which satisfies the condition of the initial state in (\ref{equ:condition_noniso_GPOMDP_B}). Here we end our proof.

\section{Identification Between Trivial and Non-Trivial Phases}
\label{sec:appendix_Identification Between Trivial and Non-Trivial Phases}

The identification of the trivial and non-trivial phases can be formulated as follows: for each episode, we determine an appropriate timestep $H_p$ such that the interval $t \in [0, H_p)$ corresponds to the non-trivial phase, and $t \in [H_p, H)$ corresponds to the trivial phase.
We define $H_p$ as the timestep at which the gripper first interacts with a target object.
However, in practice, accurately detecting this interaction is challenging, as the poses of objects are not directly observable and their shapes are unknown.
If $H_p$ is overestimated relative to the actual classification boundary $H_p^*$ (i.e., $H_p > H_p^*$), some transitions from the trivial phase will be incorrectly labeled as non-trivial, resulting in invalid data augmentations.
Conversely, if $H_p$ is underestimated ($H_p < H_p^*$), certain transitions from the non-trivial phase will be misclassified as trivial, which reduces the diversity and effectiveness of the augmentation.
To mitigate the risk of generating invalid augmentations, we conservatively choose $H_p$ to be slightly smaller than the true $H_p^*$, providing a safety margin that ensures all augmented transitions are valid, while still maximizing augmentation diversity as much as possible.

For the domain of general robot manipulation, we apply a gripper-height-based criterion for phase identification.
In this domain, the target object is initially placed on the floor, and both the gripper and the object have fixed initial heights.
The gripper’s height at any timestep can be estimated by adding the initial height to the cumulative height changes resulting from executed actions.
Therefore, we track the gripper's height to distinguish the non-trivial phase from the trivial (interaction) phase. 
Maintaining the gripper above or equal to a predefined height threshold $L_z$ prevents contact with ungrasped task objects.
At the beginning of the trajectory, when the gripper height exceeds or equals $L_z$, the corresponding transitions are identified as the non-trivial phase.
After the gripper height first drops below the threshold, indicating likely interaction with objects, all subsequent transitions are identified as the trivial phase.

For the domain of surgical grasping, we apply a timestep-based criterion for phase identification.
In this domain, the initial heights of both the target objects and the gripper are not fixed.
As a result, the gripper-height-based criterion used in general robot manipulation is not applicable, since it is difficult to reliably track the relative heights of the gripper and target object.
Instead, we define the classification timestep as $H_p = H - h_p$, where $h_p \in (0, H]$ is a predefined constant.

\bibliographystyle{IEEEtran}
\bibliography{main}

\newpage


\begin{IEEEbiography}[{\includegraphics[width=1in,height=1.25in,clip,keepaspectratio]{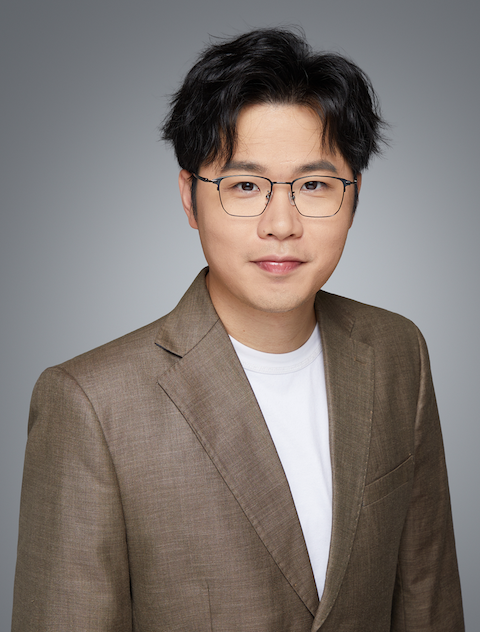}}]{Hongbin Lin}  received the
B.S. degree in mechanical and automation
engineering from Guangdong University of Technology, Guangdong, China, in 2017.
He received an M.S. degree in mechanical and automation
engineering from The Chinese University of Hong Kong, Hong Kong, China, in 2018.

He is currently a Ph.D. student in mechanical and automation
engineering from The Chinese University of Hong Kong, Hong Kong, China.
His research interests include symmetry in reinforcement learning, visuomotor learning in robot manipulation, and surgical autonomy.
\end{IEEEbiography}

\begin{IEEEbiography}[{\includegraphics[width=1in,height=1.25in,clip,keepaspectratio]{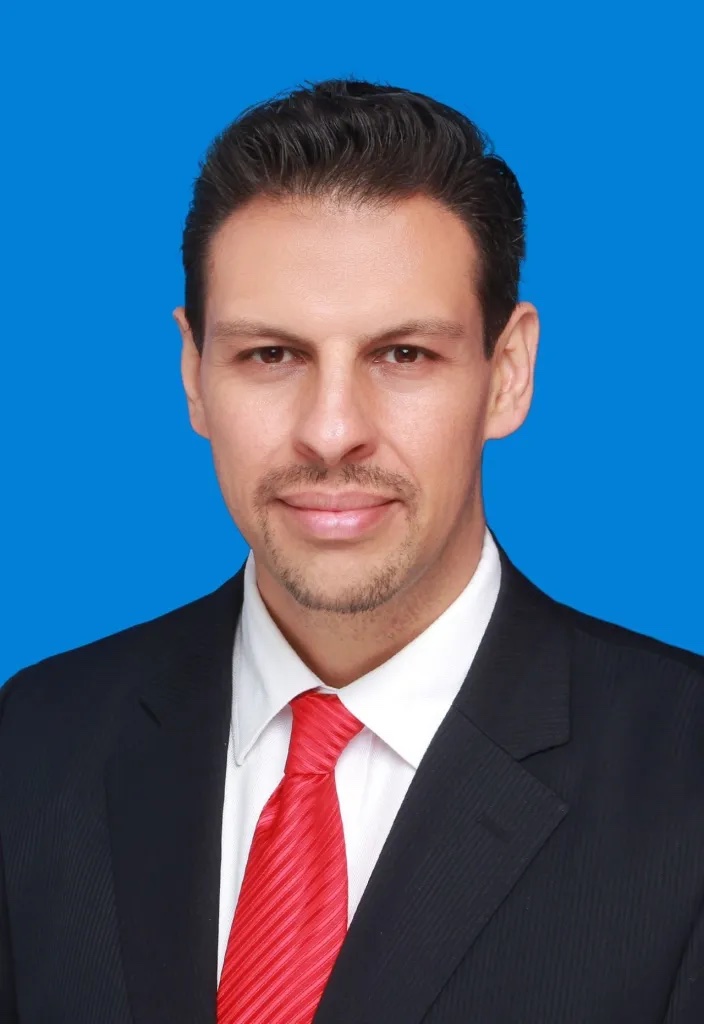}}]{Juan Rojas} (Member, IEEE) received B.S., M.S. and Ph.D. degrees in Electrical and Computer Engineering from Vanderbilt University (Nashville, TN) in May 2002, 2004, and 2009 respectively. He is currently an Assistant Professor in the Raymond B. Jones School of Engineering at Lipscomb University (Nashville, TN) and is the Director of BEST Robotics Music City. Before joining Lipscomb University, Dr. Rojas was an Assistant Research Professor in the School of Mechanical Engineering and Automation at the Chinese University of Hong Kong and Principal Investigator at the Hong Kong Center for Logistics Robotics. He also served as an Associate Professor and at the School of Electromechanical Engineering at the Guangdong University of Technology (Guangzhou, China) and an Assistant Professor in the School of Software at Sun Yat Sen University (Guangzhou, China). Dr. Rojas conducted a post‑doctoral fellowship at the Task Manipulation and Vision Center, National Institute of Advanced Industrial Science \& Technology in Tsukuba, Japan and was a Roboticist at Universal Logic Inc., Nashville, TN, USA. His research contributions include deep reinforcement learning for robotic manipulation, robot introspection, and invariant transform experience replay. Dr. Rojas has received numerous awards, including third‑place finalist in the 2024 RoboCup Autonomous Robot Manipulation Competition; Best Paper Award Finalist at the 2021 IEEE/SICE and 2019 IEEE RCAR Conferences. He was appointed IEEE Senior Member Award in 2018.
\end{IEEEbiography}

\begin{IEEEbiography}[{\includegraphics[width=1in,height=1.25in,clip,keepaspectratio]{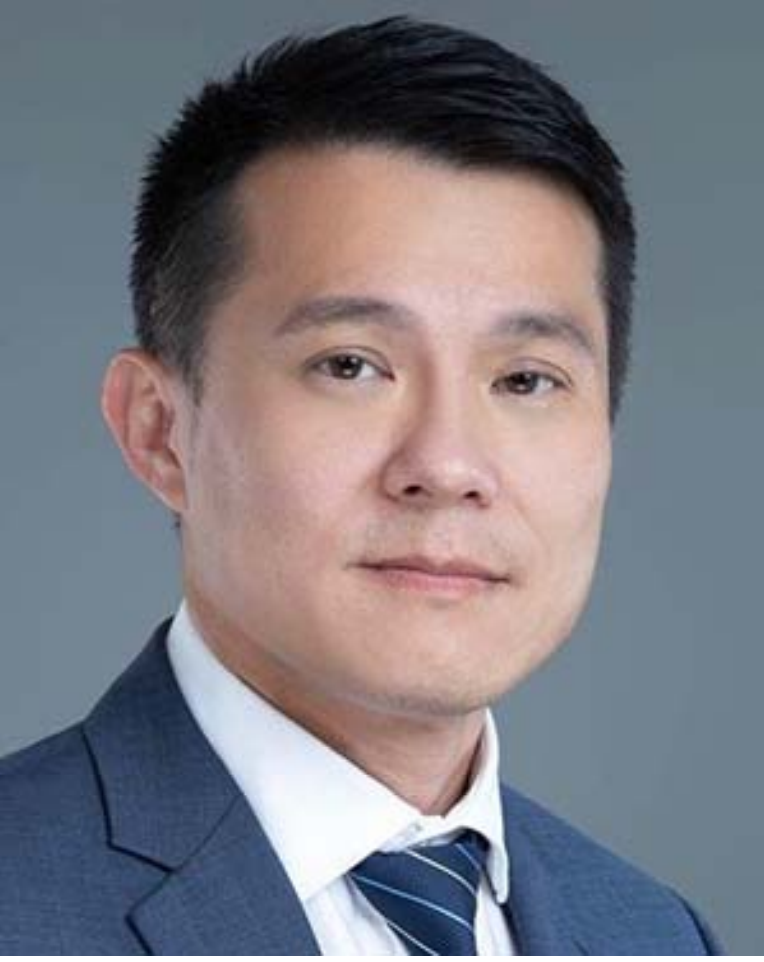}}]{Kwok Wai Samuel Au} (Member, IEEE) received the
B.S. and M.S. degrees in mechanical and automation
engineering from The Chinese University of Hong
Kong, Hong Kong, China, in 1997 and 1999, respectively, and the Ph.D. degree from the Department
of Mechanical Engineering, Massachusetts Institute
of Technology (MIT), Cambridge, MA, USA, in
2007.

He is currently a Professor with the Department of
Mechanical and Automation Engineering, The Chinese University of Hong Kong and also the Director
with the Multi-Scale Medical Robotics Center, InnoHK. Before joining The
Chinese University of Hong Kong, he was the Manager of Systems Analysis
with the Department of New Product Development, Intuitive Surgical, Inc.,
Sunnyvale, CA, USA. He coinvented the FDA-approved da Vinci Single-Site
surgical platform and was also a founding team member for the da Vinci ION
system. During his Ph.D. study, he also coinvented the MIT Powered Ankle-foot
Prosthesis with Prof. H. Herr. He holds ten U.S. Patents and more than eight
pending U.S. Patents.

Dr. Au was the recipient of numerous awards including the first prize in
the American Society of Mechanical Engineers Student Mechanism Design
Competition in 2007, Intuitive Surgical Problem Solving Award in 2010, and
Intuitive Surgical Inventor Award in 2011.
\end{IEEEbiography}

\vspace{11pt}

\vfill

\end{document}